\documentclass[conference]{IEEEtran}
\usepackage[numbers]{natbib}
\usepackage{amsmath}

\usepackage{listings}
\usepackage{url}

\usepackage[pdftex]{graphicx}

\usepackage{amssymb}

\usepackage{stfloats}

\begin{document}
%
\title{Robust Unsupervised Transient Detection\\With Invariant Representation based on\\the Scattering Network}

\author{\IEEEauthorblockN{Randall Balestriero}
\IEEEauthorblockA{Ecole Normale Superieure\\
Paris, France\\
Email: randallbalestriero@gmail.com}
\and
\IEEEauthorblockN{Behnaam Aazhang}
\IEEEauthorblockA{Rice University\\
Houston, TX\\
Email: aaz@rice.edu}
}

\maketitle

\begin{abstract}
We present a sparse and invariant representation with low asymptotic complexity for robust unsupervised transient and onset zone detection in noisy environments. This unsupervised approach is based on wavelet transforms and leverages the scattering network from Mallat et al. by deriving frequency invariance. This frequency invariance is a key concept to enforce robust representations of transients in presence of possible frequency shifts and perturbations occurring in the original signal. Implementation details as well as complexity analysis are provided in addition of the theoretical framework and the invariance properties. In this work, our primary application consists of predicting the onset of seizure in epileptic patients from subdural recordings as well as detecting inter-ictal spikes.
\end{abstract}

\IEEEpeerreviewmaketitle

\section{Motivations}

\begin{figure}[t!]
\begin{center}
\includegraphics[width=3.5in]{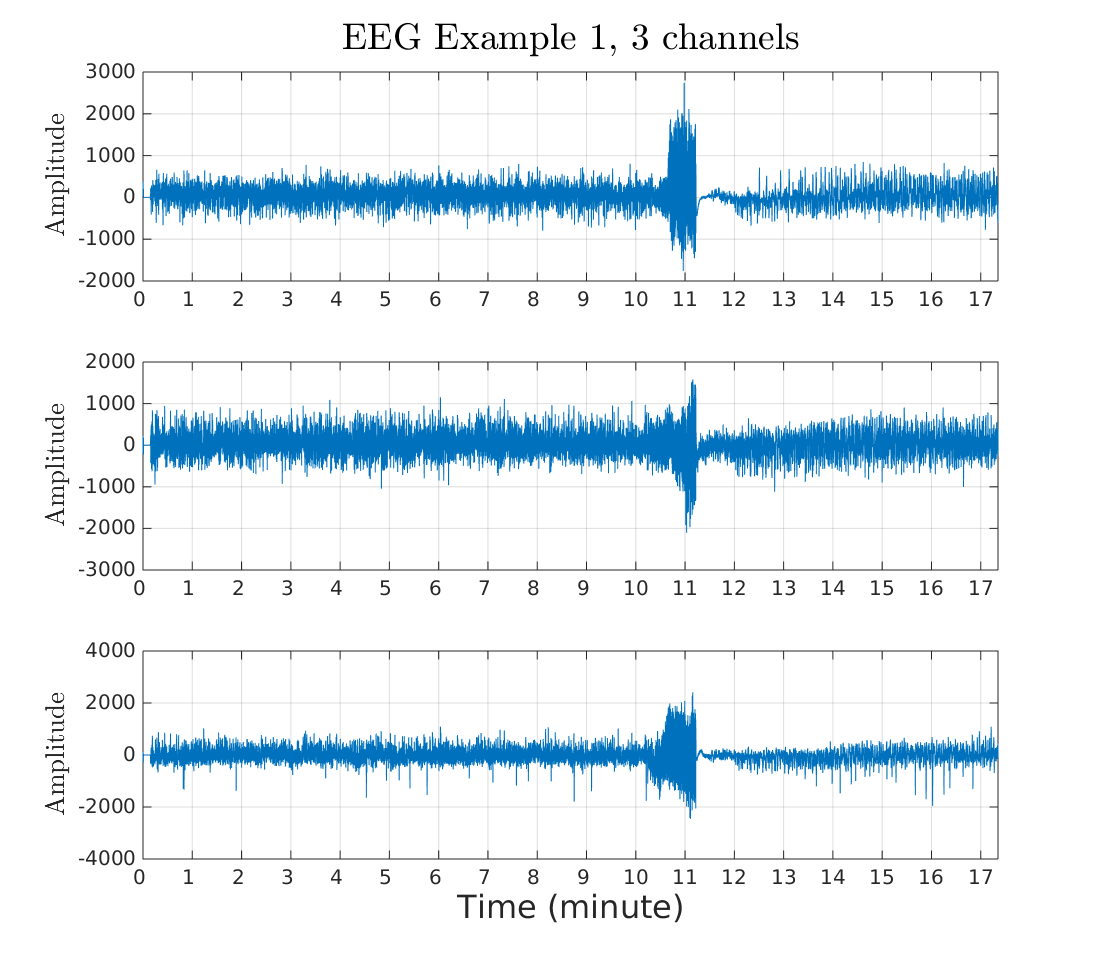}\caption{Three different channels showing the pre-ictal, ictal and post-ictal phase for a patient.}\label{EXAMPLE123}
\end{center}
\end{figure}

\begin{figure}[t!]
\begin{center}
\includegraphics[width=3.5in]{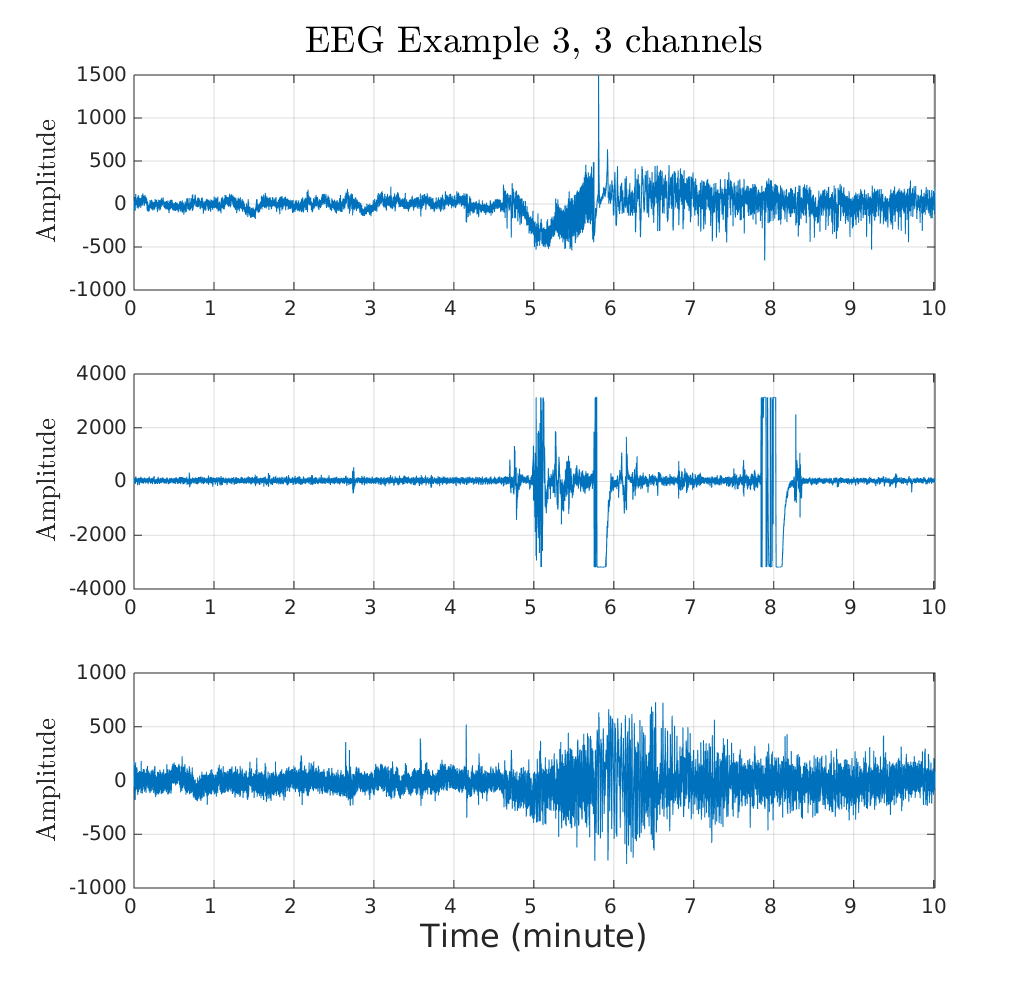}\caption{Three channels on another patient during seizure where the middle plot depicts discontinuities due to human movements and are thus artefacts induced by the happening seizure.}\label{EXAMPLE1234}
\end{center}
\end{figure}

\begin{figure}[t!]
\begin{center}
\includegraphics[width=3.5in]{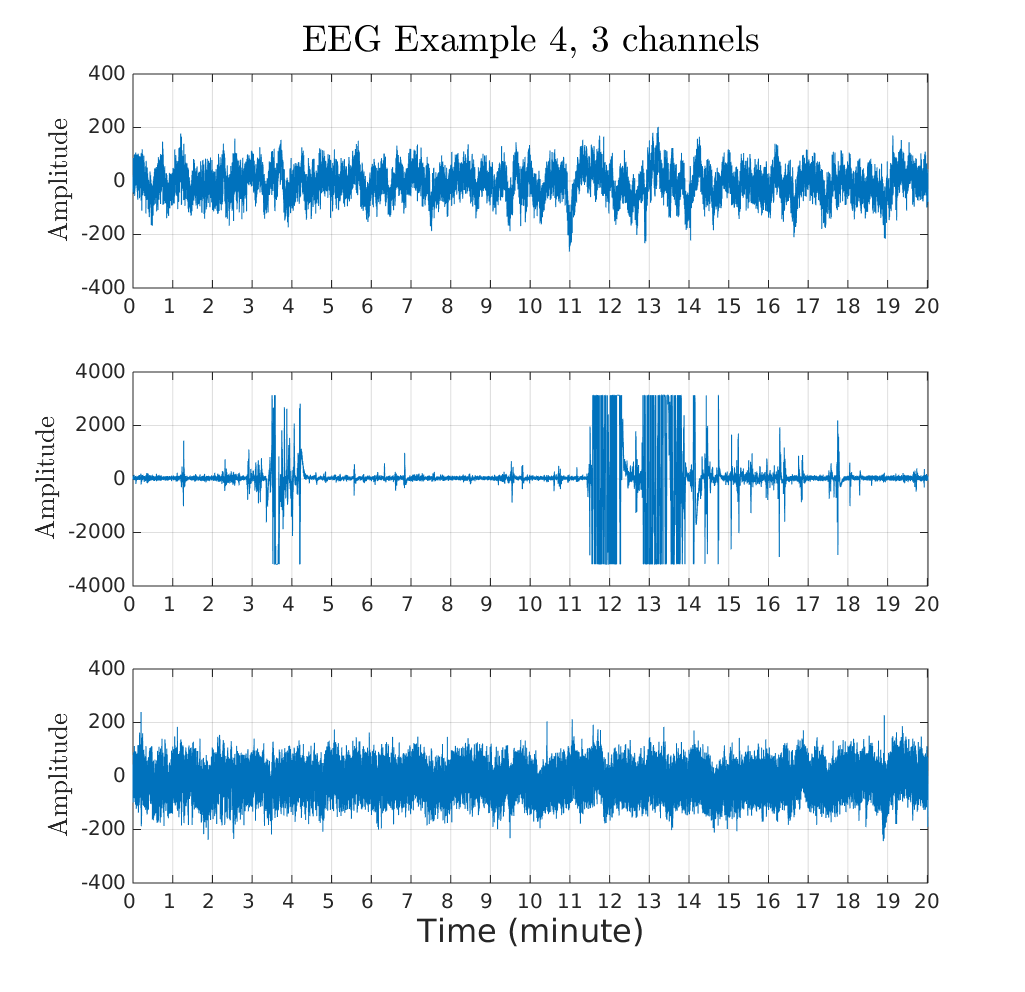}\caption{Three channels during a normal brain activity, no seizure happening, yet the middle plot depicts discontinuities implied by natural human movements.}\label{EXAMPLE12345}
\end{center}
\end{figure}

In multiple contexts, and in particular with electroencephalography (EEG) signals, it is necessary to have an unsupervised pre-processing technique to highlight relevant features in noisy environments. In fact, to obtain an intra-cranial EEG (iEEG) signal, electrodes are implanted underneath the dura and in some cases deep inside brain tissues in order to record the average electrical activity over a small brain region. This results in unbalanced and noisy datasets of small sizes putting supervised approaches aside. When considering the task of seizure prediction, the key unsupervised representation should be sparse and parsimonious \cite{meyer1993wavelets} but most importantly invariant to seizure independent events we denote as "noise" for our task \cite{oyallon2015deep}\cite{sifre2013rotation}. This last property will be the key to ensure the robustness of the technique since it implies stability of the transient detection despite the presence of frequency-shift perturbations. For iEEG data of epileptic patients, we are dealing with signals such as presented in Fig. \ref{EXAMPLE123},\ref{EXAMPLE1234} and \ref{EXAMPLE12345}.
In this context, we wish to highlight transients i.e. local bursts of energy embedded in ambient but not necessarily white noise generated through a possibly nonstationary process. These breaking points in the intrinsic dynamical structure of the signals are thus our features of interests, the ones we aim to capture. Biologically, they represent inter-ictal spikes which are often sign of an upcoming seizure or other types of neuro-features s.a. high-frequency oscillations (HFO), pre-ictal activity.
The complete absence of learning of the scattering network leads to robust representations even with only few observations. It has also been shown that representations using the scattering network are very efficient when dealing with stationary processes as those encountered in music genre classification \cite{chen2013music} and texture classification \cite{bruna2013invariant} due to the global time invariance property. This ability to well capture the ambient signal is thus the key of our approach which, simply put, will extract transients by comparing the signal and its scattering representation.
\section{Scattering Network}

The scattering network is a signal processing architecture with multiple processing steps. It is based on cascades of linear transforms and nonlinear operations on an input signal $x$. Commonly, this is achieved by successive wavelet transforms followed by the application of a complex modulus \cite{anden2014deep}. One can notice that several approaches in machine learning aim to bring linear separability between the features of interest through the application of many linear and nonlinear operators usually followed by dimensionality reduction aiming to bring local invariance. This is exactly what the scattering network does to its input $x$. Moreover, this framework can be seen as a restricted case of a Convolutional Neural Network \cite{lecun1995convolutional} where the filters are fixed wavelets, the activation function is the complex modulus and the translation invariance is obtained through time averaging.

\subsection{Preliminaries}
In order to perform a wavelet transform, we first create the bank of filters used to decompose the signal. A wavelet filter-bank results from dilation and contraction of a mother wavelet $\psi_0$. This mother wavelet, by definition, satisfies the following admissibility condition:
\begin{equation}
\int_{\mathbb{R}}\psi_0(t) dt =0 \iff \hat{\psi}_0(0)=0,
\end{equation}
where we define by $\hat{\psi}_0$ the Fourier Transform of $\psi_0$ as
\begin{equation}
\hat{\psi}_0(\omega)=\int_\mathbb{R}\psi_0(t)e^{-i \omega t} dt.
\end{equation}
The dilations are done using scales factors $\lambda \in \Lambda$  following a geometric progression parametrized by the number of wavelets per octave $Q$ and the number of octave to decompose, $J$. This is defined explicitly as the set of scale factors
\begin{equation}
\Lambda= \{ 2^{1+j/Q},j=0,...,J\times Q-1\}.
\end{equation}\label{deflambda}
We can now express our filter-bank as a collection of dilated version of $\psi_0$ by the scale $\lambda$ as
\begin{equation}
\psi_{\lambda}(t)=\frac{1}{\lambda}\psi_0\left( \frac{t}{\lambda}\right)
\; \forall \lambda \in \Lambda,
\end{equation}
or in the Fourier domain with
\begin{equation}
\hat{\psi}_{\lambda}(\omega)=\lambda \hat{\psi}_0(\lambda \omega)
\; \forall \lambda \in \Lambda.
\end{equation}

Wavelets are defined as band-pass filters with constant ratio of width to center frequency and they are localized in both the physical and the Fourier domain \citep{mallat1999wavelet}.
Since all the children wavelets are scaled version of the mother wavelet, they also integrate to $0$. Therefore, even with an infinite number of wavelets, none will be able to capture the low frequencies, the mean, in the limit. A complementary filter is thus introduced, namely, the scaling function $\phi$. In order to capture these reaming low frequencies, this filter satisfies the following criteria:
\begin{equation}
\int_\mathbb{R}\phi(t) dt=1 \iff \hat{\phi}(0)=1.
\end{equation}
The basis used to transform $x$ is thus the collection $\{  \phi, \psi_\Lambda  \}$ made of all the band-pass filters as well as the scaling function, the latter giving rise to the scattering coefficients per say.

\subsection{Scattering Decomposition}
\begin{figure}[t!]
\begin{center}
\includegraphics[width=3.5in]{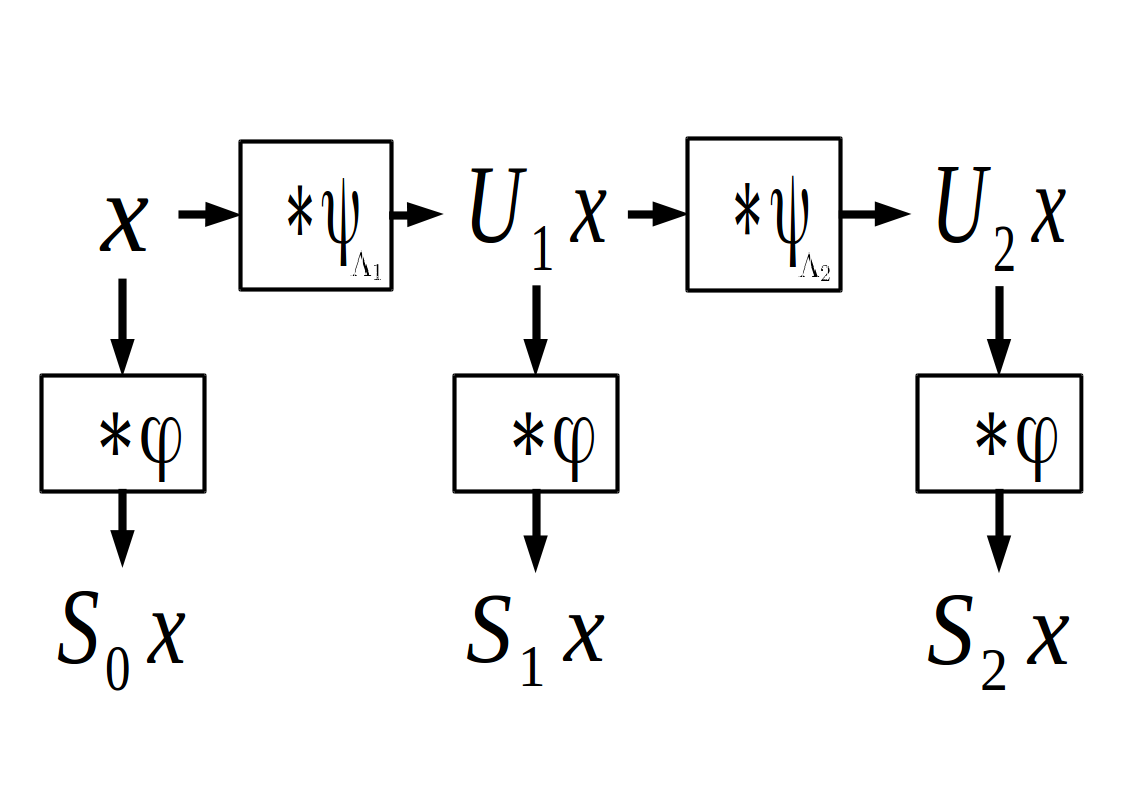}\caption{Scattering network architecture leading to $S_2x$ which we used in building our representation.}\label{scattt}
\end{center}
\end{figure}

We now present the scheme used to decompose a signal through multiple layers of the scattering networks via multiple bank-filters each one being specific to each layer. Since we will now have successions of wavelet transforms, we define by $\Lambda_i$ the $i^{th}$ set of scales as
\begin{equation}
\Lambda_i= \{ 2^{1+j/Q_i},j=0,...,J_i\times Q_i-1\},
\end{equation}
for each of the layer dependent hyper-parameters $J_i,Q_i$. We now describe formally the scattering network operations summarized in Fig. \ref{scattt} up to the second layer, where transients are encoded, as shown in \cite{anden2014deep}.

The decomposition of the given signal $x$ with all the band-pass filters $\psi$ is now defined with the $U$ operator recursively for all the layers where $*$ denotes the convolution operator as
\begin{equation}
\begin{matrix}
U_0x=x,\\
U_ix:=\left\{|U_{i-1}x * \psi_{\lambda}|,\;\forall \lambda \in \Lambda_i\right\}  .
\end{matrix}
\end{equation}
This operator applies all the band-pass filters to the output of the previous operator starting from the input signal. For example, $U_1x$ and $U_2x$ are defined explicitly as:
\begin{equation}
\begin{matrix}
U_1x(t,\lambda_1)= \left| \left| \int x(\tau) \psi_{\lambda_1}(t -\tau) d\tau \right|\right|,\\
\forall \lambda_1 \in \Lambda_1
\end{matrix}
\end{equation}
and
\begin{equation}
\begin{matrix}
U_2x(t,\lambda_1,\lambda_2)=
\left| \left| \int \left( \int x(\tau) \psi_{\lambda_1}(\xi -\tau) d\tau \right) \psi_{\lambda_2}(t-\xi) d \xi \right| \right|,\\
\forall \lambda_1 \in \Lambda_1,\forall \lambda_2 \in \Lambda_2
\end{matrix}
\end{equation}
One can notice that $U_1x$ coefficients correspond to the scalogram, a standard time-scale representation.
We define similarly the scattering operators as
\begin{equation}
\begin{matrix}
S_0x:=x*\phi ,\\
S_ix:=U_{i}x * \phi,
\end{matrix}
\end{equation}
which brings time invariance by smoothing its input via the scaling function $\phi$. In term of dimensionality,  for the $i^{th}$ layer, $U_ix$ and $S_ix$ depend on the parameters $(t,\lambda_1,...,\lambda_i),\lambda_1 \in \Lambda_1,...,\lambda_i\in\Lambda_i$ and thus are of index-dimension $i+1$.


We now describe our representation which is based on the scattering coefficients obtained from $S_2x$ to end up with a representation suited for transient detection in noisy environments.
\section{Sparse Representation for Transient Detection}

\begin{figure}[t!]
\begin{center}
\includegraphics[width=3.5in]{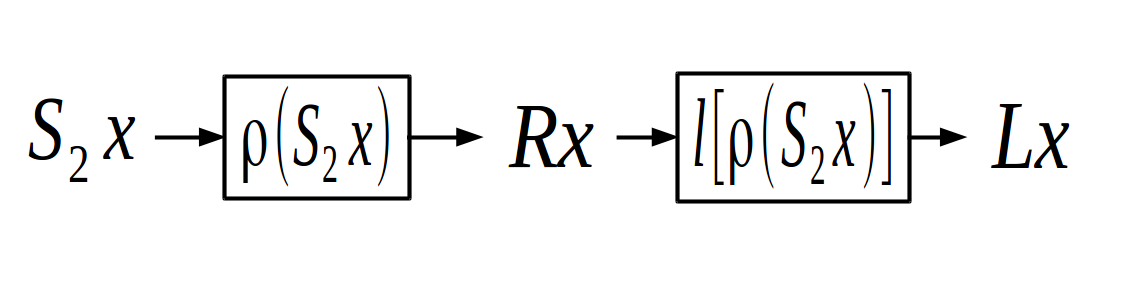}\caption{From the second layer of the scattering network $S_2x$, a threshold $\rho$ is applied leading to $Rx$, followed by local dimensionnality reduction to obtain our final representation $Lx$}\label{scattt222}
\end{center}
\end{figure}

Using the scattering network coefficients $S_2x$, two main transformations are applied leading to our final representation $Lx$, defined below. The first operator $\rho(S_2x)$ is the application of a thresholding followed by a nonlinearity improving the SNR by highlighting transients while collapsing to $0$ the ambient background texture. The second transformation defined as $l\left(\rho(S_2x)\right)$, brings frequency invariance through application of local dimensionality reduction. These two stages will lead to the new $2$D representation
with index-dimension $t$ and $\lambda_2$
\begin{equation}
S_2x \rightarrow \rho(S_2x):=Rx \rightarrow l\left(\rho(S_2x)\right):=Lx,
\end{equation}
also presented in Fig. \ref{scattt222}. We now describe each of these two transformations in detail.

\subsection{Nonlinear Thresholding Through Scattering Coefficients}
The scattering coefficients $S_2x(t,\lambda_1,\lambda_2)$ are capturing the transients diluted in the ambient signal activity. In order to emphasize transient structures, a Lipschitz nonlinearity denoted as  $\rho$ is applied to detect bursts of energy \cite{allard2014expansive}. In fact, we need to weight the $S_2x$ representation such that outliers, i.e transient structures, are emphasized from the ambient signal. By considering the ambient signal as a centroid, transients become outliers and thus a point-wise comparison of the coefficients and the centroids will measure the intensities of these events leading to $Rx$.

\begin{figure}[t!]
\begin{center}
\includegraphics[width=3.5in]{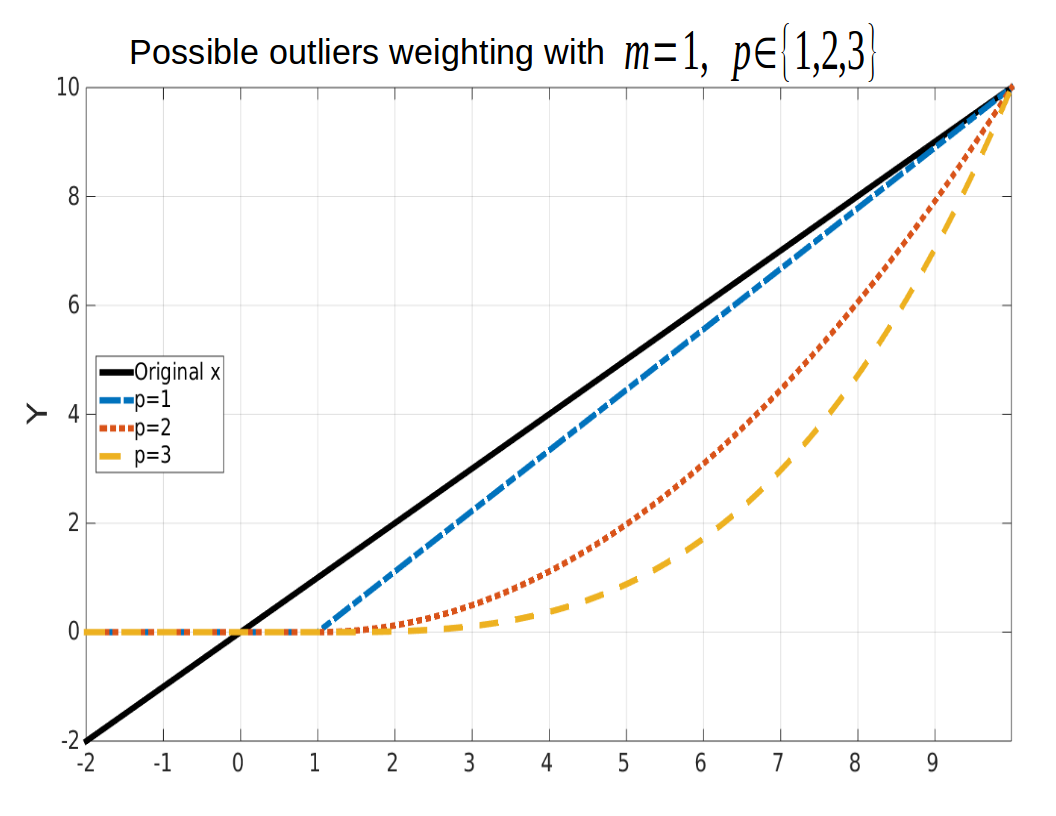}\caption{Example of different expansive nonlinearities with threshold at $1$ applied to the black line.}\label{sparsity04}
\end{center}
\end{figure}

We define the $\rho$ operator which will include the thresholding and the polynomial nonlinearity as 
\begin{equation}
\rho^{(p)}_m(x)=\left\{ \begin{matrix}
0 \text{ if $x \leq m $ }\\
(x-m)^p \text{ if $x > m$ }
\end{matrix} \right. .
\end{equation}
When applied to time-series, the output will also be normalized so that the array before and after the application of $\rho$ will have the same infinite-norm. The nonlinearity $\rho^{(p)}_m$ is depicted in Fig. \ref{sparsity04} for different values of $p$ and $m=1$.
The $p$ parameter can be seen as the $p-$norm distance used after application of the threshold to weight the coefficients based on their distances from the given threshold value $m$. With great $p$, only clear outliers are kept whereas with $p<1$, the operator $\rho$ is a compressive nonlinearity. In our approach, the threshold value $m$ will represent the centroid of each scale and is defined as the median in order to provide a robust estimate of the ambient signal energy. We thus define the threshold value $m$ as
\begin{equation}  
m(\lambda_1,\lambda_2):=median_t \; S_2x(t,\lambda_1,\lambda_2), \forall (\lambda_1,\lambda_2) \in \Lambda_1 \times \Lambda_2.
\end{equation}
The threshold value is time-invariant but not constant over $\lambda_1$ nor $\lambda_2$ leading to level specific thresholding with independent estimations of the ambient signal centroids. As a result, our first stage transform $Rx$ is defined in the following way
\begin{align}
Rx(t,\lambda_1,\lambda_2):=&\rho^{(p)}_{m(\lambda_1,\lambda_2)}\left( S_2x\left( t,\lambda_1,\lambda_2\right) \right),\nonumber \\ &\forall (\lambda_1,\lambda_2) \in \Lambda_1 \times \Lambda_2 
\end{align}

\begin{figure}[t!]
\begin{center}
\includegraphics[width=2.5in]{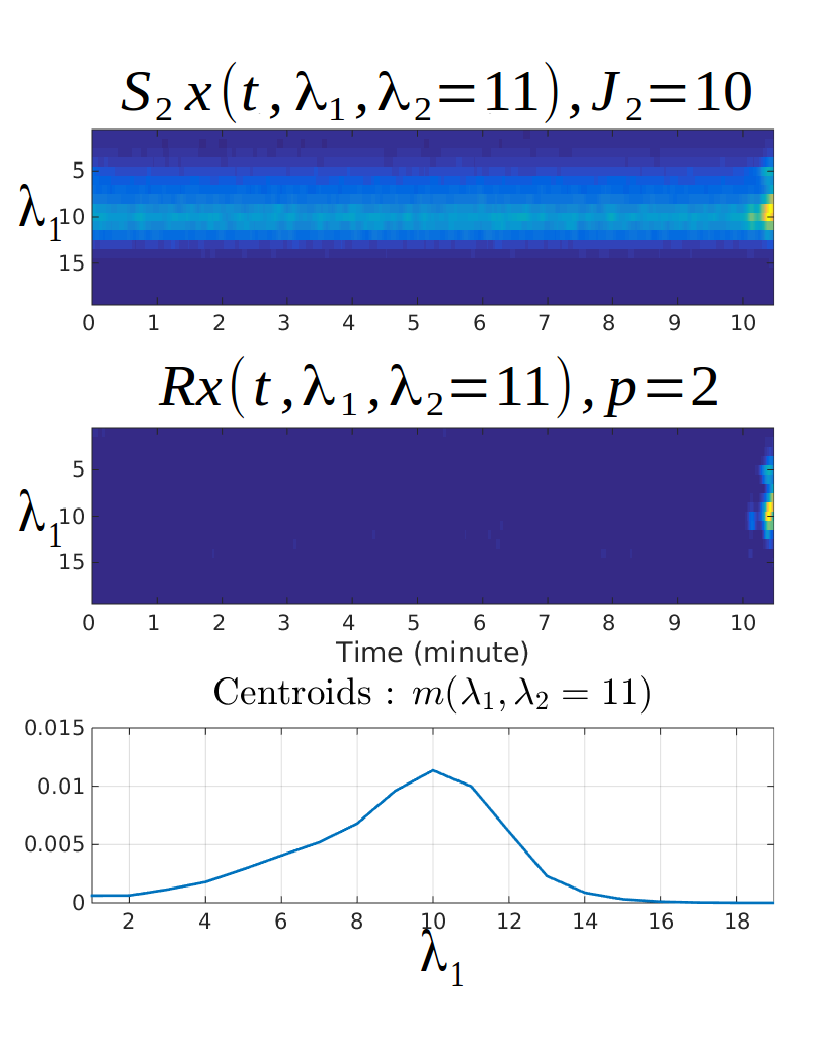}\caption{Top: $S_2x(t,\lambda_1,\lambda_2=11)$ coefficients with $\lambda_2$ fixed to be the $11^{th}$ filter. Middle: $Rx(t,\lambda_1,\lambda_2=11)$ representation  highlighting the background to foreground thresholding increasing the SNR. Bottom: the median (centroid) for each scale used as the threshold value. The input signal is a iEEG recording with a seizure starting at the end.}\label{sparsity33}
\end{center}
\end{figure}

The use of the median ensures that we reach a sparsity of at least $50\%$ since by definition this threshold will have as many coefficients below it than above it. However, this also means that the transients we are trying to detect should not be present in more than $50\%$ of the time otherwise the less energetic ones will be cut off. We thus obtain a sparse representation, illustrated by Fig. \ref{sparsity33}. This new representation is able to capture the nonstationary structure while removing the ambient noise leading to a higher SNR.
In fact, because of the compressive property of the nonlinearity for small coefficients, the values close to the threshold will collapse to $0$. 
In all our experiment, setting $p=2$ led to efficient weighting parameter to achieve our sparse representation.

\subsection{Local Dimensionality Reduction}

Transients are defined as bursts of energy being localized in time with a broad active frequency spectrum \cite{randall2}. A time-frequency representation of a lone transient will lead to a Dirac function in the time domain versus an almost uniform representation in the frequency domain.
Since by construction $Rx$ captures transient structures, this resulting representation is highly redundant over $\lambda_1$ due to the structure of transients as shown in Fig. \ref{sparsity33}. In fact, when fixing $\lambda_2=k$, the 2D representation $Rx(t,\lambda_1,k)$ is sparse in time and redundant over the $\lambda_1$ frequency dimension. In other word, because of their intrinsic structures, events captured in $Rx$ are oversampled in the $\lambda_1$ dimension.
As a result, we are able to reduce this redundancy by aggregating, combining, the $\lambda_1$ dimension of $Rx$ with respect to each  $\lambda_2$. In order to do so, two general dimensionality reduction techniques are studied: PCA and max-pooling.
PCA performs an optimized linear dimensionality reduction 
over all $\lambda_1$ for each $\lambda_2$. This leads to a linear and time-invariant dimensionality reduction. 
In order to do this, we compute the covariance matrix of $Rx(t,\lambda_1,\lambda_2)$ with respect to the $\lambda_1$ dimension for each $\lambda_2$ fixed. This results in $|\Lambda_2|$ matrices of size $|\Lambda_1| \times |\Lambda_1|$.  By mean of the spectral theorem \cite{reed1908methods} we diagonalize these matrices, and obtain their spectrum. The eigenvector corresponding to the highest eigenvalue is used to project our data onto the direction with most variability\cite{PCA} for each of the matrices leading to the new representation only indexed on $t$ and $\lambda_2$.
We define the projection achieved by the PCA of $Rx$ on the corresponding eigenvector as follows: 
\begin{equation}
Lx(t,\lambda_2)=
\text{PCA}_{\lambda_1}\left[Rx(t,\lambda_1,\lambda_2)\right] ,
\end{equation}
In order to provide a nonconstant dimensionality reduction scheme over time, we also introduce the max-pooling over all the $\lambda_1$ for each $\lambda_2$ leading to nonlinear dimensionality reduction. Note that max-pooling is a standard nonlinear dimensionality reduction technique widely used in neural networks \cite{giusti2013fast}\cite{nagi2011max} and is defined here as:
\begin{equation}
Lx(t,\lambda_2)=\max_{\lambda_1}(Rx(t,\lambda_1,\lambda_2)).
\end{equation}

\begin{figure}[h]
\begin{center}
\includegraphics[width=2.5in]{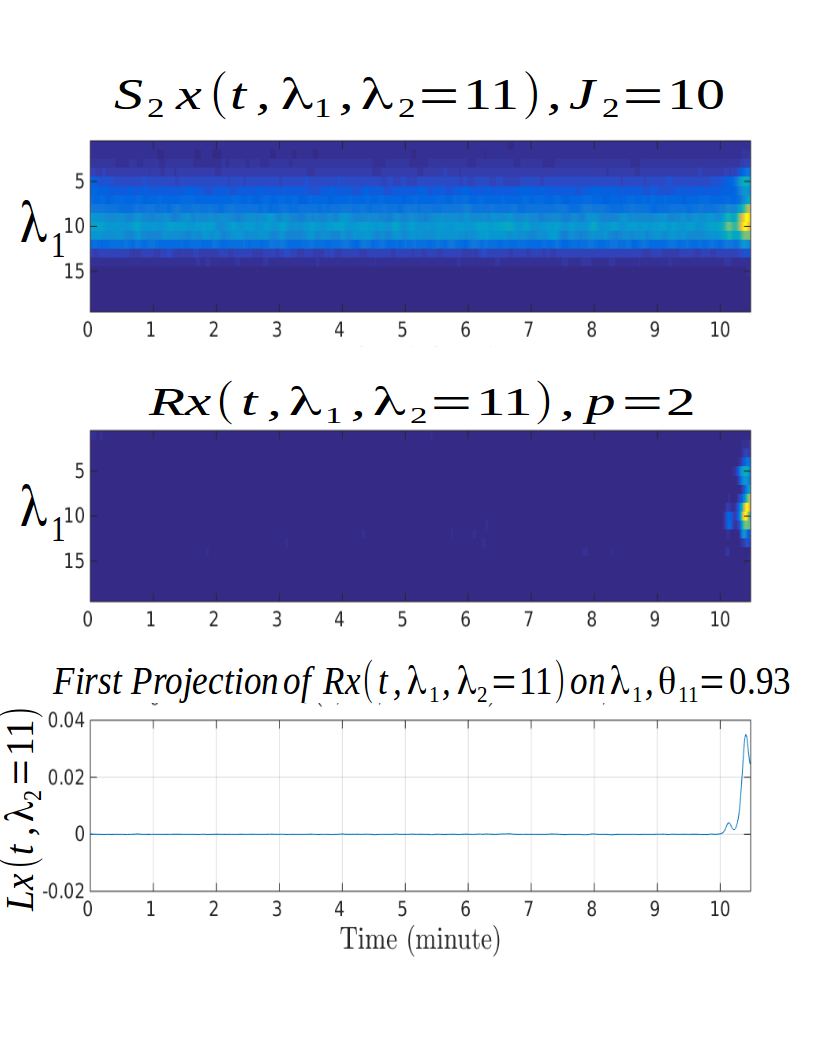}\caption{Top: $S_2x(t,\lambda_1,\lambda_2=11)$ coefficients with $\lambda_2$ fixed to be the $11^{th}$ filter. Middle : $Rx(t,\lambda_1,\lambda_2=11)$ representation  highlighting the background to foreground thresholding increasing the SNR. Bottom : aggregation of information over $\lambda_1$ using PCA for a the fixed $\lambda_2$ leading to $Lx(t,\lambda_2=11)$}\label{sparsity015}
\end{center}
\end{figure}

Since we applied one PCA for each $\lambda_2$ and kept each first projected vector, we can evaluate the projections by analyzing the normalized eigenvalues. In order to clarify the following notations, we now denote by $\theta_{\lambda_2}$ the first eigenvalue for each of the performed PCA. 
Considering variance as a feature of interest is not always justified. However, since in our case PCA is applied on $Rx$, variance results from the present transient structures. And since $\theta_{\lambda_2}$ represents the amount of variance retained by the first eigenvector, it follows that greater $\theta_{k}$ implies presence of nonstationary structures in $R(t,\lambda_1,k)$ with no frequency modulations i.e. transients.
\begin{figure}[t!]
\begin{center}
\includegraphics[width=3.1in]{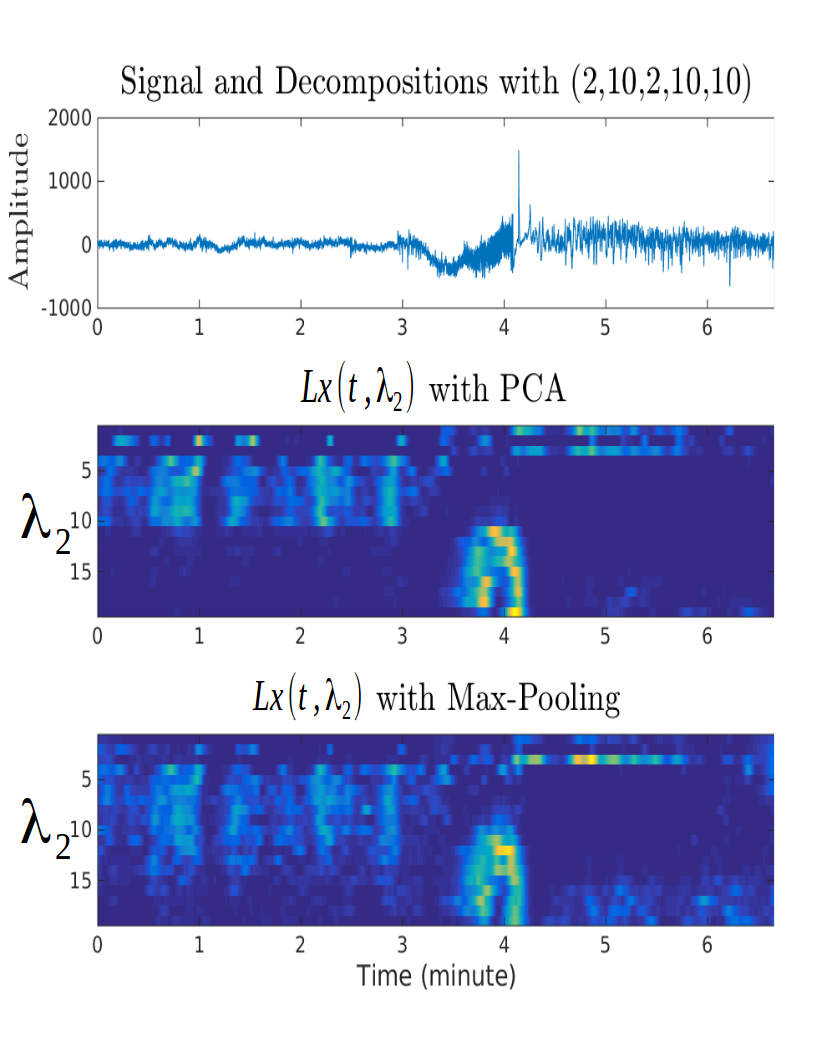}\caption{Top: EEG signal example with a seizure starting at about $3$ min. Middle: $Lx$ representation computed using local PCA. Bottom: $Lx$ representation computed using local max-pooling.}\label{sparsity121}
\end{center}
\end{figure}
In Fig. \ref{sparsity121} the use of PCA reveals inter-ictal activities (0-3 min) without interfering with the pre-ictal (3-3:30 min) and ictal detection (3:30-4:10 min) whereas max-pooling makes these coefficients become negligible.
\begin{figure}[t!]
\begin{center}
\includegraphics[width=2.8in]{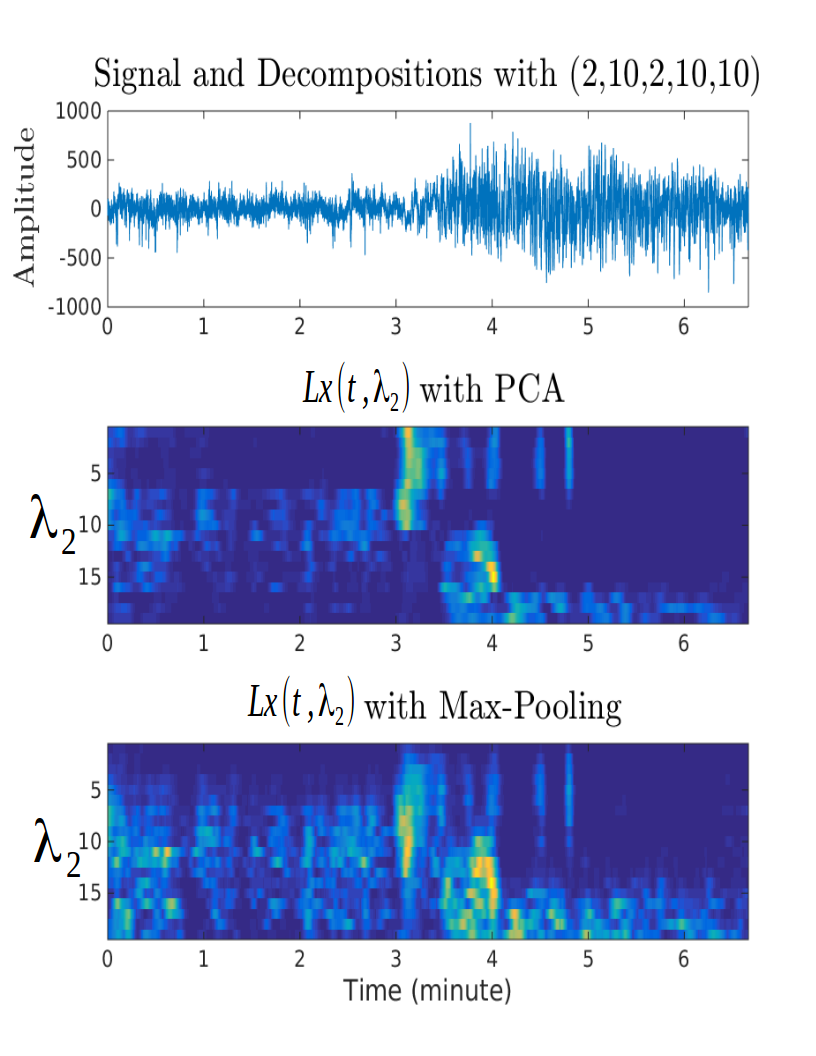}\caption{Top: noisy EEG signal example with a seizure starting at about $3$ min. Middle : $Lx$ representation computed using local PCA. Bottom : $Lx$ representation computed using local max-pooling.}\label{sparsity131}
\end{center}
\end{figure}
\begin{figure}[t!]
\begin{center}
\includegraphics[width=2.8in]{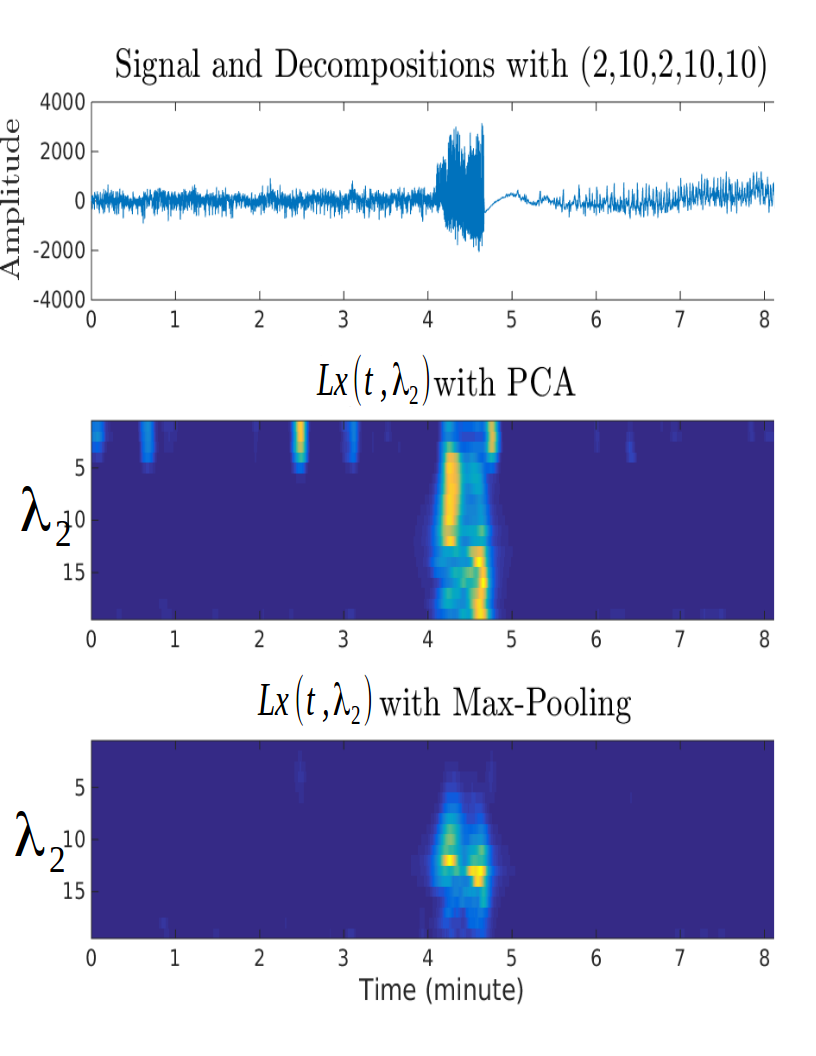}\caption{Top: EEG signal example with a seizure starting at about $r$ min and showing inter-ictal spikes prior to it. Middle : $Lx$ representation computed using local PCA. Bottom : $Lx$ representation computed using local max-pooling.}\label{sparsity132}
\end{center}
\end{figure}
In Fig. \ref{sparsity131}, both dimensionality reduction algorithms are able to catch different events but sharper transitions appear when using PCA. For our application, the case presented in Fig. \ref{sparsity132} is particularly interesting since the PCA allows to clearly observe that the tempo at which the transients occur is decreasing as the seizure happens  \cite{quiroga2002frequency}.

Thus, the main difference between max-pooling and PCA results from the fact that for max-pooling, the pooled coefficient for each point in time $t_1$ is computed independently from past or future points $t_2 \not = t_1$. In fact, the computation of $Lx(t_1,\lambda_2)$ is independent from the computation $Lx(t_2,\lambda_2), \forall \; t_1 \neq t_2$.
On the contrary, for PCA, the linear combination leading to $Lx(t,\lambda_2)=\sum_{i=1}^{J_1Q_1}\alpha_iRx(t,i,\lambda_2)$ uses the same weighting coefficients $\alpha_i$ for all the time domain and treat each of these arrays entirely when computing the best projection as opposed to a Mixture of Probabilistic PCA approach \cite{tipping1999probabilistic}. 

\subsection{Feature Vector}
Since the aim of this approach is to provide an unsupervised efficient feature extraction algorithm, there is still a need to use the obtained representation in a machine learning algorithm either for unsupervised clustering with a k-NN as will be done below or for a supervised classification task with a SVM for example \cite{cortes1995support}. As a result, we now present what are the obtained features from our approach as well as their respective dimensionality.
We denote by $V$ the feature vector composed with the standard scattering coefficients ($S_0x(t)$, $S_1x(t,\lambda_1)$, $S_2x(t,\lambda_1,\lambda_2)$) , the ambient signal levels $m(\lambda_1,\lambda_2)$, the first eigenvalue of each PCA done for each $\lambda_2$ on $Rx(t,\lambda_1,\lambda_2)$ and finally the corresponding projections $Lx(t,\lambda_2)$.
The total dimension disregarding the time-index is thus $O(Q_1J_1+Q_2J_2+Q_1J_1Q_2J_2)$ and exactly:
\begin{equation}
dim(V)=1+J_1Q_1+2J_1Q_1J_2Q_2+2J_2Q_2.
\end{equation}
For the EEG cases shown here, the baseline parameters are $(J_1,Q_1,J_2,Q_2)=(2,10,2,10)$ leading to a feature vector of size $861$. Note that this size can be reduced by using the max-pooling approach thus dropping the $\theta_{\lambda_2}$ coefficients.

\subsection{Computational Complexity}
In addition of providing a tractable feature vector, this approach is also suited for large scale problems. We detail here the asymptotic complexity of each step that is performed in addition of using an efficient implementation presented in \cite{balestriero2015scattering}.
The scattering transform is of complexity $O(n \log(n))$ with $n$ being the size of the input signal. 
The median computation is $O(n)$ when using the selection algorithm and the fact that the median is the $\frac{n}{2}^{th}$ order statistic. 
Finally, PCA is $O(p^2n+p^3)$ where $n$ is the size of the input signal and here $p=\lambda_1$ where for our examples we have  $\lambda_1=20$. 
But for extreme cases, the PCA could be replaced by the max-pooling.
We now present in the next two sections the invariance properties present in $Lx$ as well as some possible uses of the eigenvalues $\theta_{\lambda_2}$ as robust descriptors of the different behaviors of the studied signals in order to perform onset zone detection and transient detection. In short, local change in the underlying generative process of the observed signals while moving through time.

\subsection{Invariants}
By definition, the suggested representation $Lx$ based on $S_2x$ will inherit the local time-invariance property from the scattering network but will also provide a global frequency invariance over the $\lambda_1$ dimension. In fact, if we rearrange the $\lambda_1$ dimension in $U_1x$ the resulting $Lx$ will not change with either of the two dimensionality reduction techniques: PCA or max-pooling.
Frequency invariance is important since we seek a sparse representation of transients. We thus seek stability with respect to changes in the $\lambda_1$ dimension. 
We will prove for the max-pooling case that the representation is the same when applied to $U_1x$ or $\tilde{U}_1x$ where $\tilde{U}_1x$ is a random permutation of the $\lambda_1$ of $U_1x$, the proof for the PCA case is similar with the reordering of the covariance matrix. Thus it will include the simple frequency translation case as well as any kind of bijection s.a. reordering. The random permutation is denoted by $\sigma$.

\begin{IEEEproof}
By considering $U_1x'(t,\lambda_1):= U_1x(t,\sigma(\lambda_1))$ we can derive $S_2x'(t,\lambda_1,\lambda_2):=S_2x(t,\sigma(\lambda_1),\lambda_2)$. We also require $\sigma$ to be bijective.
\begin{align*}
Lx'(t,\lambda_2) &=\max_{i \in \lambda_1} \rho^p_{median_tS'_2(t,i,\lambda_2)} \left( S'_2(t,i,\lambda_2) \right) \\
&=\max_{i \in \lambda_1}\rho^p_{median_tS_2(t,\sigma(i),\lambda_2)} \left( S_2(t,\sigma(i),\lambda_2) \right) \\
&=\max_{i \in \sigma^{-1}(\lambda_1)}\rho^p_{median_t S_2 (t,i,\lambda_2)} \left( S_2(t,i,\lambda_2) \right) \\
&=\max_{i \in \lambda_1}\rho^p_{ median_t S_2 (t,i,\lambda_2)} \left( S_2 (t,i,\lambda_2) \right) \\
&=Lx (t,\lambda_2)
\end{align*}
\end{IEEEproof}
Note that the dimensionality reduction can be done over parts of the $\lambda_1$ dimension to only bring local frequency invariance instead of global frequency invariance.

\subsection{Analysis of the Eigenvalues}
\begin{figure}[t!]
\begin{center}
\includegraphics[width=3.2in]{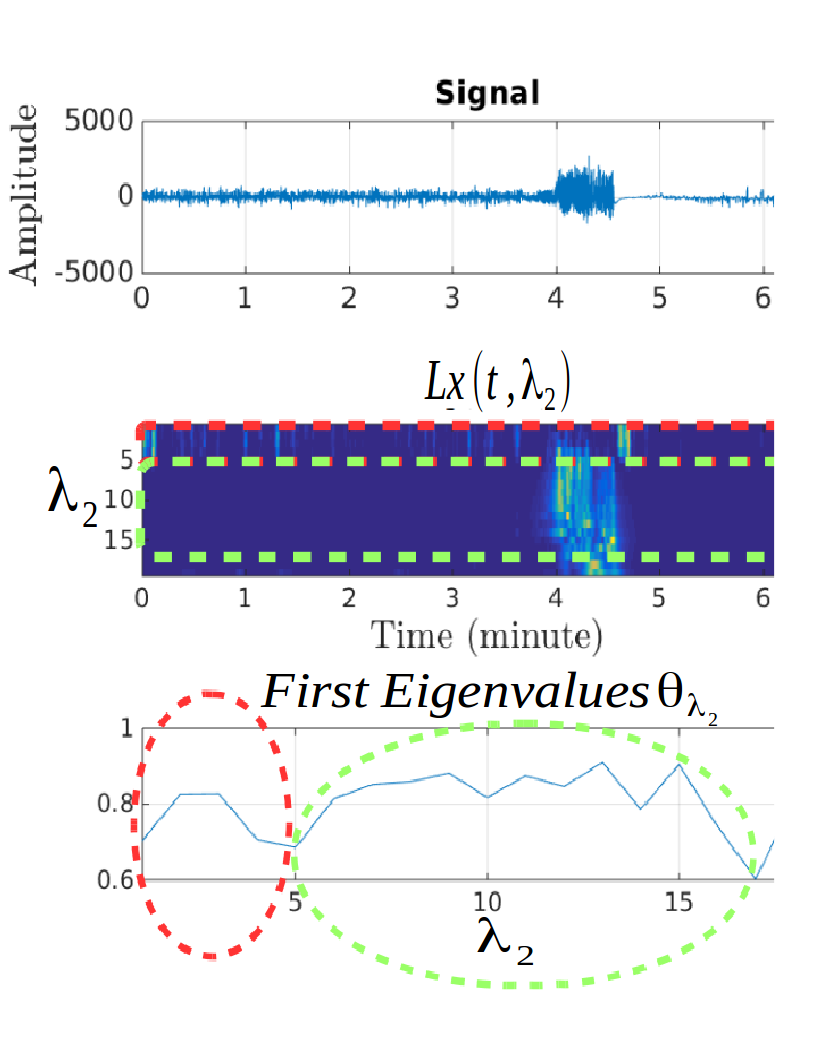}\caption{Top: Seizure happening at time 4-4:40 min. Middle: $Lx$ representation showing two types of events, inter-ictal spikes and ictal activities. Bottom: Clustering of the first eigenvalues of each applied PCA.}\label{sparsitye1}
\end{center}
\end{figure}

The distribution of the eigenvalues of the signal of interest over $\lambda_2$ also contain meaningful information on the presence of transients for each frequency. In fact, if for some $\lambda_2$, the representation $Rx(t,\lambda_1,\lambda_2)$ contains primarily background residuals and no transients, then, the corresponding PCA will be ineffective at reducing this representation to one vector and thus the corresponding first eigenvalue will be small relative to other eigenvalues of different $\lambda_2$ containing transients.
For the case in Fig. \ref{sparsitye1}, it is in fact possible to see two main clusters which have been circled. The separation between these two clusters is easily seen through the succession of small and high eigenvalues.
The first cluster $\lambda_2=0-5,$ is correlated with interictal activities (0-3:50 min) whereas the second cluster $\lambda_2=6-17$ is related to pre-ictal and ictal events (3:50:4:45 min).
\begin{figure}[t!]
\begin{center}
\includegraphics[width=3.3in]{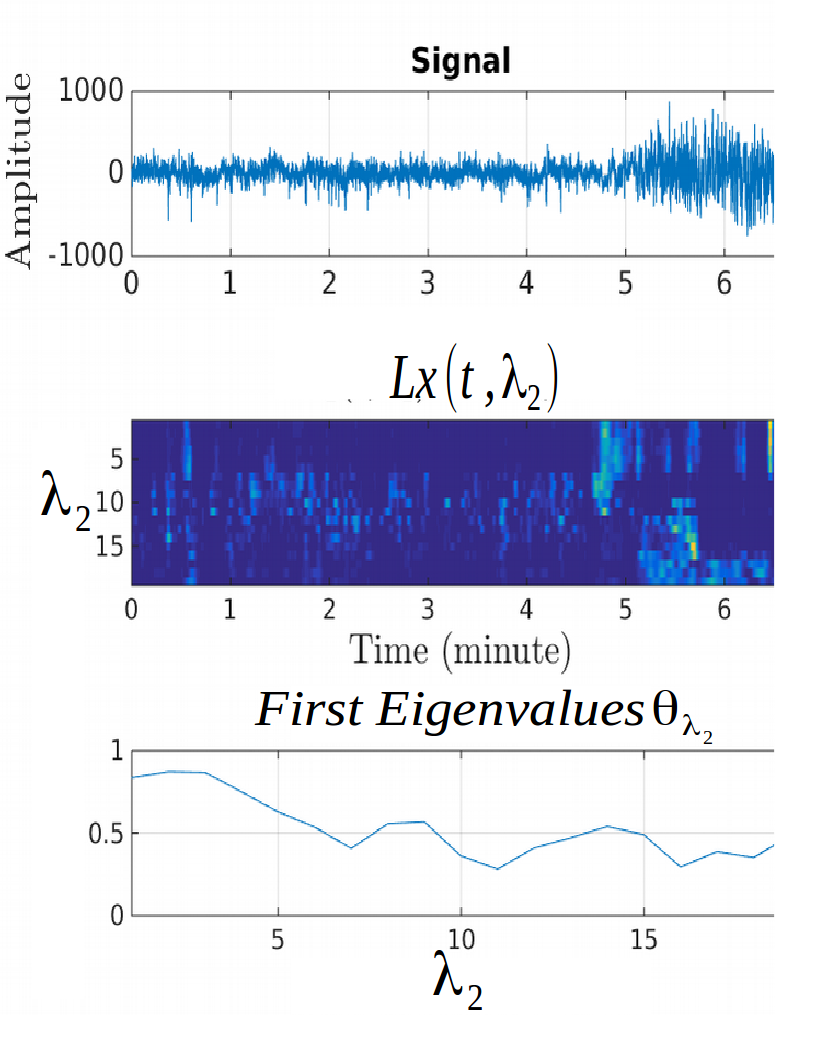}\caption{Top: Seizure activities starting at 5min. Middle: Representation showing the start of ictal activities prior to the 5min seizure start.}\label{sparsitye2}
\end{center}
\end{figure}
In the second example presented in Fig.\ref{sparsitye2} one can see a first cluster $\lambda_2=0-7$ related to High Frequency Oscillations \cite{guirgis2013role} (HFO) whereas the second one $\lambda_2=8-11$ is linked to inter-ictal activities. The last two ones $\lambda_2=11-17$ are related to pre-ictal and ictal activities.

\begin{figure}[t!]
\begin{center}
\includegraphics[width=3.5in]{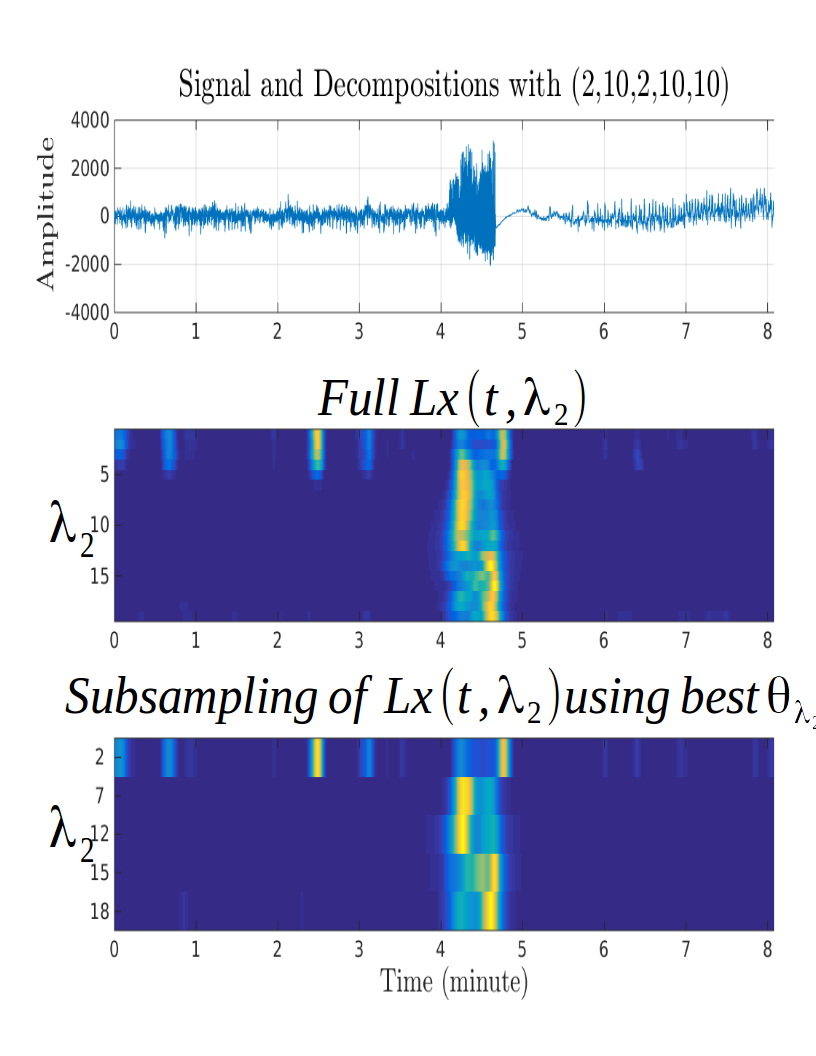}\caption{Example where we only keep the best representatives based on $\theta_{\lambda_2}$ reducing by a factor of $3$}\label{sparsitye3}
\end{center}
\end{figure}

\begin{figure}[t!]
\begin{center}
\includegraphics[width=3.1in]{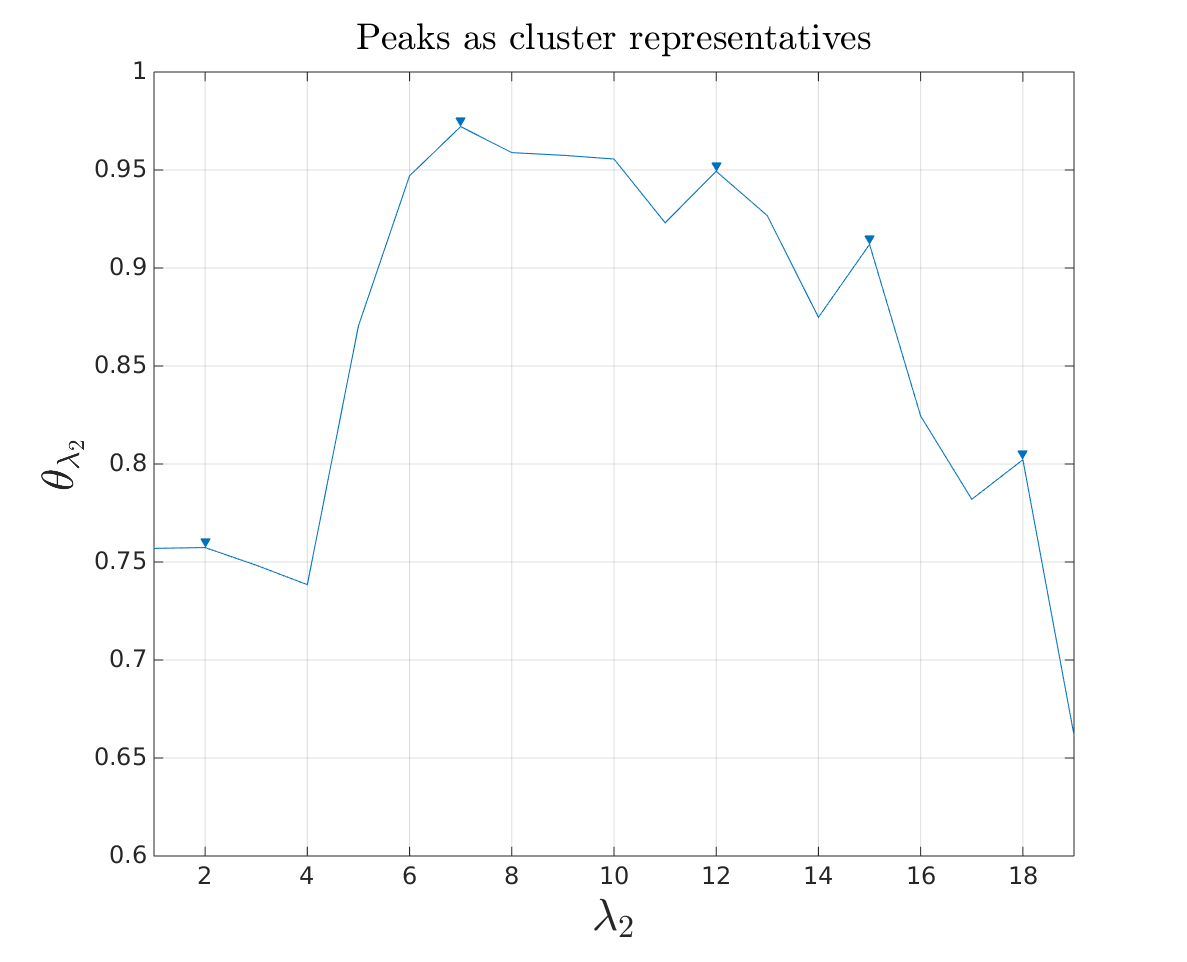}\caption{Top: Seizure happening at time 4-4:40 min. Middle: Full representation showing the inter-ictal activities, and the changes happening during the seizure activities. Bottomn: Subsample representation only kipping the best $\lambda_2$ according to the best eigenvalues.}\label{sparsitye4}
\end{center}
\end{figure}

Since we have some structure in the distribution of the $\theta_{\lambda_2}$ we can use only the "representatives" of each cluster as seen in Fig. \ref{sparsitye3} and Fig. \ref{sparsitye4} allowing us to reduce the number of $\lambda_2$ coefficients according to the eigenvalue cluster centroids. This results in a truncated  representation $Lx(t,\lambda_2)$ where only the $\lambda_2$ corresponding to the best $\theta_{\lambda_2}$ are kept as depicted in Fig. \ref{sparsitye3}. The best $\theta_{\lambda_2}$ are taken to be the local maximum or the cluster best representative as shown in Fig. \ref{sparsitye4}.
A soft version of this $\lambda_2$ sub-sampling can be done through a weighting of the $Lx$ representation with respect to their corresponding eigenvalues leading to
\begin{equation}
Lx(t,\lambda_2=i)=Lx(t,\lambda_2=i) \times \theta_{\lambda_2=i}.
\end{equation}
This a posteriori weighting aims to emphasize the projections containing the more transients by again using the eigenvalues as qualitative measures.
\section{Application on iEEG Data}
\subsection{EEG Data Presentation}
EEG data are recordings of neural activities through electrodes. In our cases these electrodes were intra-cranial and with a frequency sampling of $1000$Hz. These signals might contain important artifacts created by sharp moves of the patients.
Concerning the parameters, we selected $(J_1,Q_1,J_2,Q_2)=(2,10,2,10)$ and $p=2$. 
The first four parameters are standard in the literature for EEG analysis $J_1$ should lead to a high pass decomposition from the Nyquist frequency to $1$Hz. We also used $Q_1=Q_2=2$ wavelets per octave to have a more precise and redundant representation in the frequency domain without sacrificing computation time. Then the $J_2$ parameter is taken the same as $J_1$. The $p=2$ parameter is taken so that sparsity is increased and we effectively have a nonlinearity leading to the best result in term of increasing the SNR. A coefficient $p\leq 1$ improves the noise energy while $p\geq 3$ is too discriminant only highlighting the event of best energy in the signal.



\subsection{Real-Time Seizure Prediction}

A challenging problem in signal processing deals with iEEG data specifically for the task of seizure prediction. This challenge is crucial in order to develop electro-stimulation devices with the goal to treat epileptic patients. We thus use the developed framework for this task since the frequency invariance property as well as the design of the $Lx$ representation are suited for detecting the kind of change in the underlying generating process of the observed signal one aims to detect. In order to perform our approach in a real-time prediction setting, we will apply it on overlapping chunks of the signal. The idea is to show a presentation over the whole signal but at point $t$ only information of the past is used. We chose a window size of $60 000$ bins corresponding to $1$ minute of signal given our frequency sampling of $1$ kHz and with a large overlapping of $58$s in order to model changes happening in a $2$s window. Since we don't have yet labeled data to measure our performances, it will be done qualitatively and visually.
\begin{figure}[t!]
\begin{center}
\includegraphics[width=3.5in]{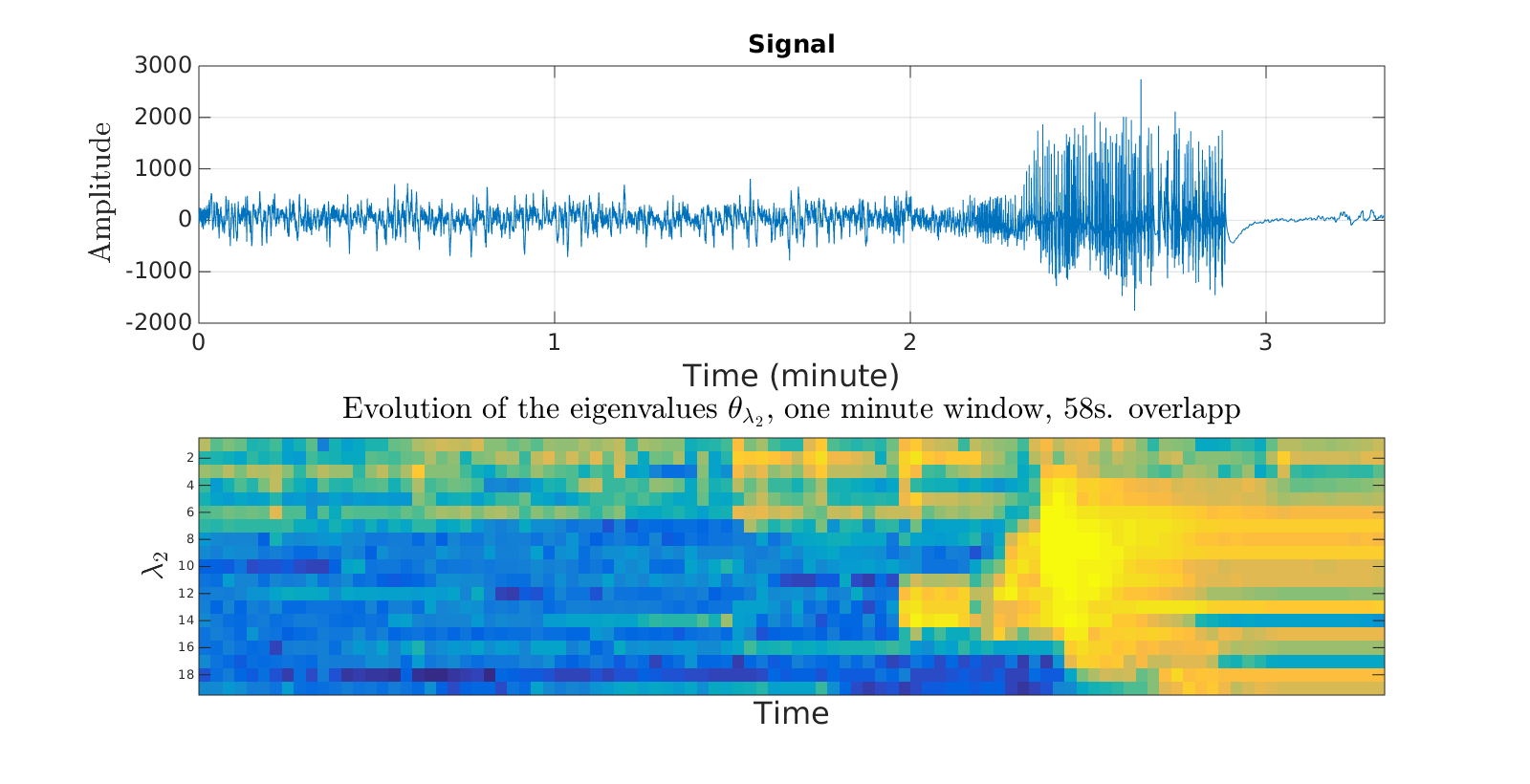}\caption{Evolution of the distribution of the eigenvalues over time with pre-ictal activities emphasized at time 2-2:20min and ictal activities at time 2:20-2:50min.}\label{sparsity1}
\end{center}
\end{figure}

\begin{figure}[t!]
\begin{center}
\includegraphics[width=3.5in]{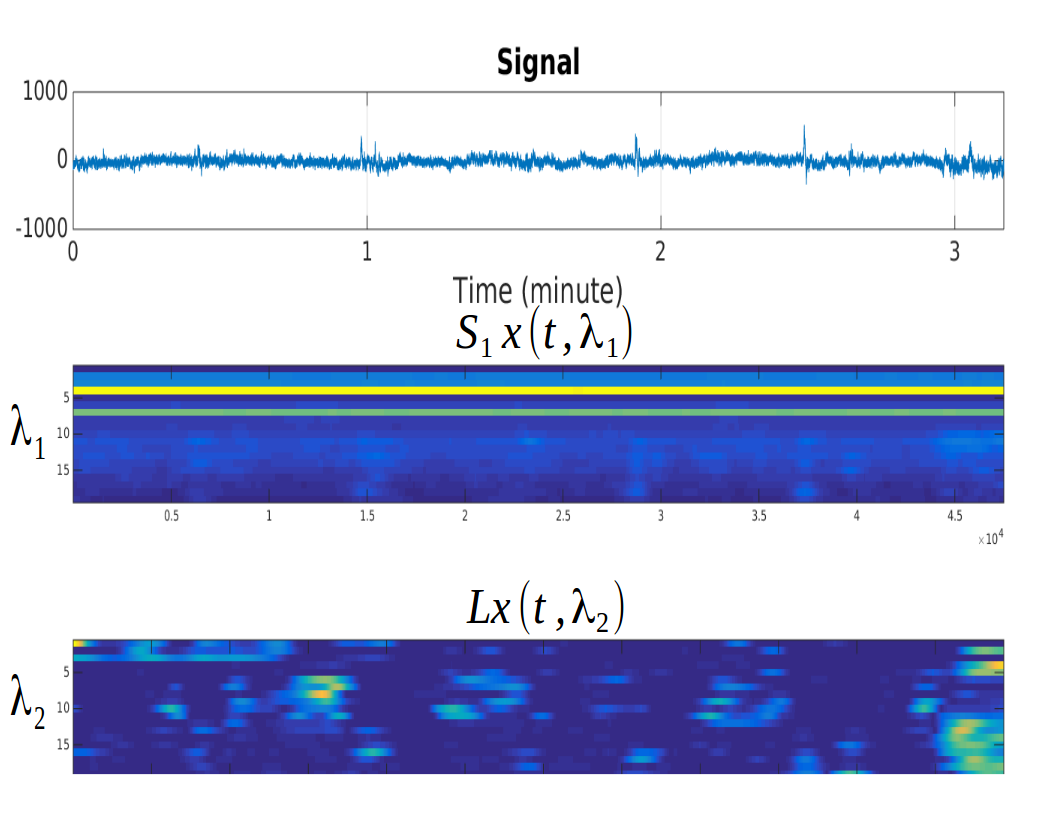}\caption{Using $Lx$ to be sure of the inter ictal phase. Clear change in the medium to low frequencies.}\label{sparsity5}
\end{center}
\end{figure}

In Fig. \ref{sparsity1}, the state transition is clear from inter-ictal ($0$-$2$ min) to pre-ictal ($2$-$2:20$ min) if we consider the medium to low frequencies. In fact, the high-frequency part is always active since inter-ictal spikes are always present. In addition we also distinguish the pre-ictal state from the ictal state ($2:20$-$2:55$ min) clearly.
\begin{figure}[h]
\begin{center}
\includegraphics[width=3.5in]{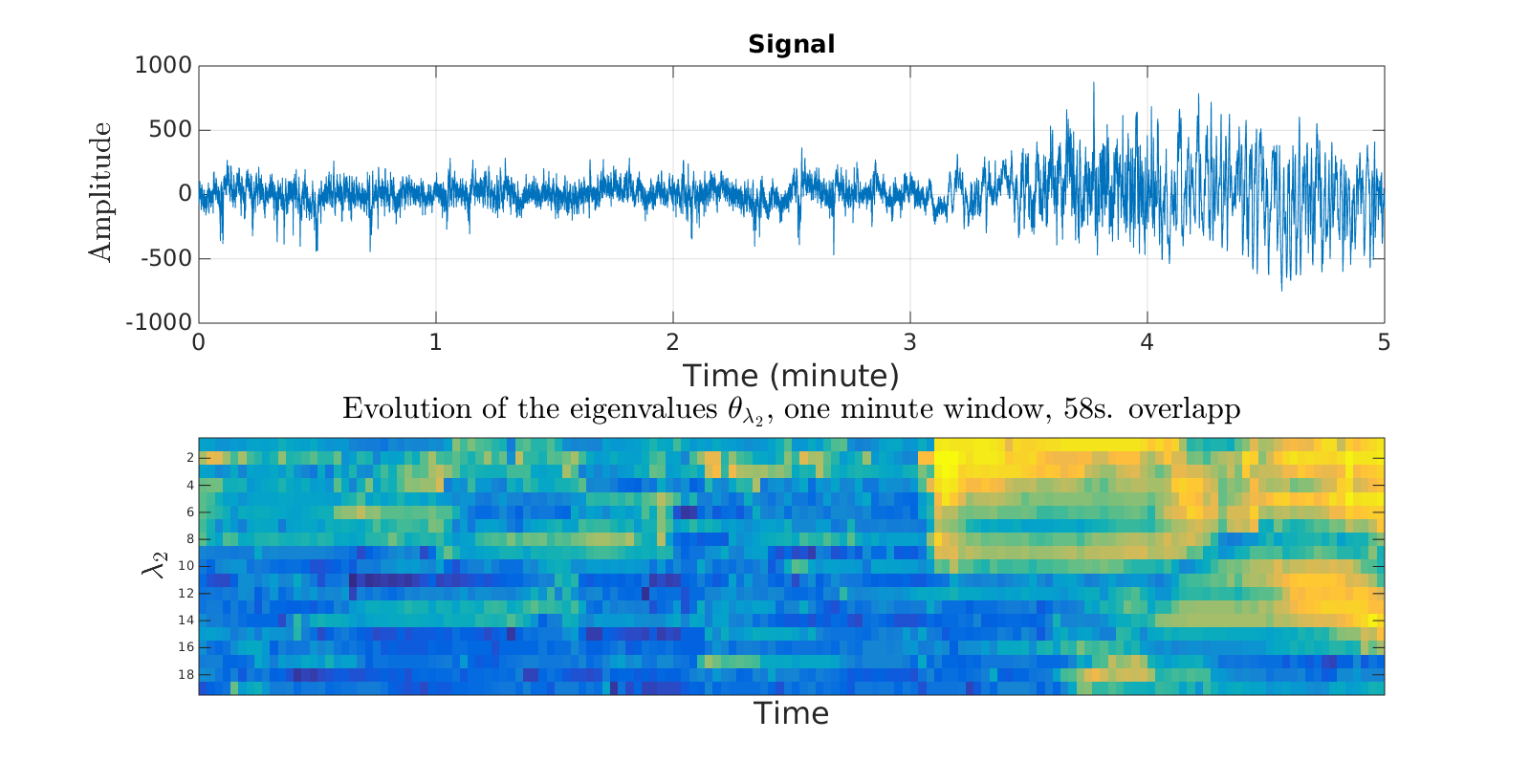}\caption{Evolution of the distribution of the eigenvalues over time showing the start of the ictal activities at 3min.}\label{sparsity2}
\end{center}
\end{figure}
In the second case presented in Fig. \ref{sparsity2} the state transition is sharp and it is interesting to note that it appears ahead in time of the seizure highlighting the pre-ictal state ($3$-$3:30$ min). This kind of representation can easily be used for prediction with a simple energy threshold for example.
\begin{figure}[t!]
\begin{center}
\includegraphics[width=3.5in]{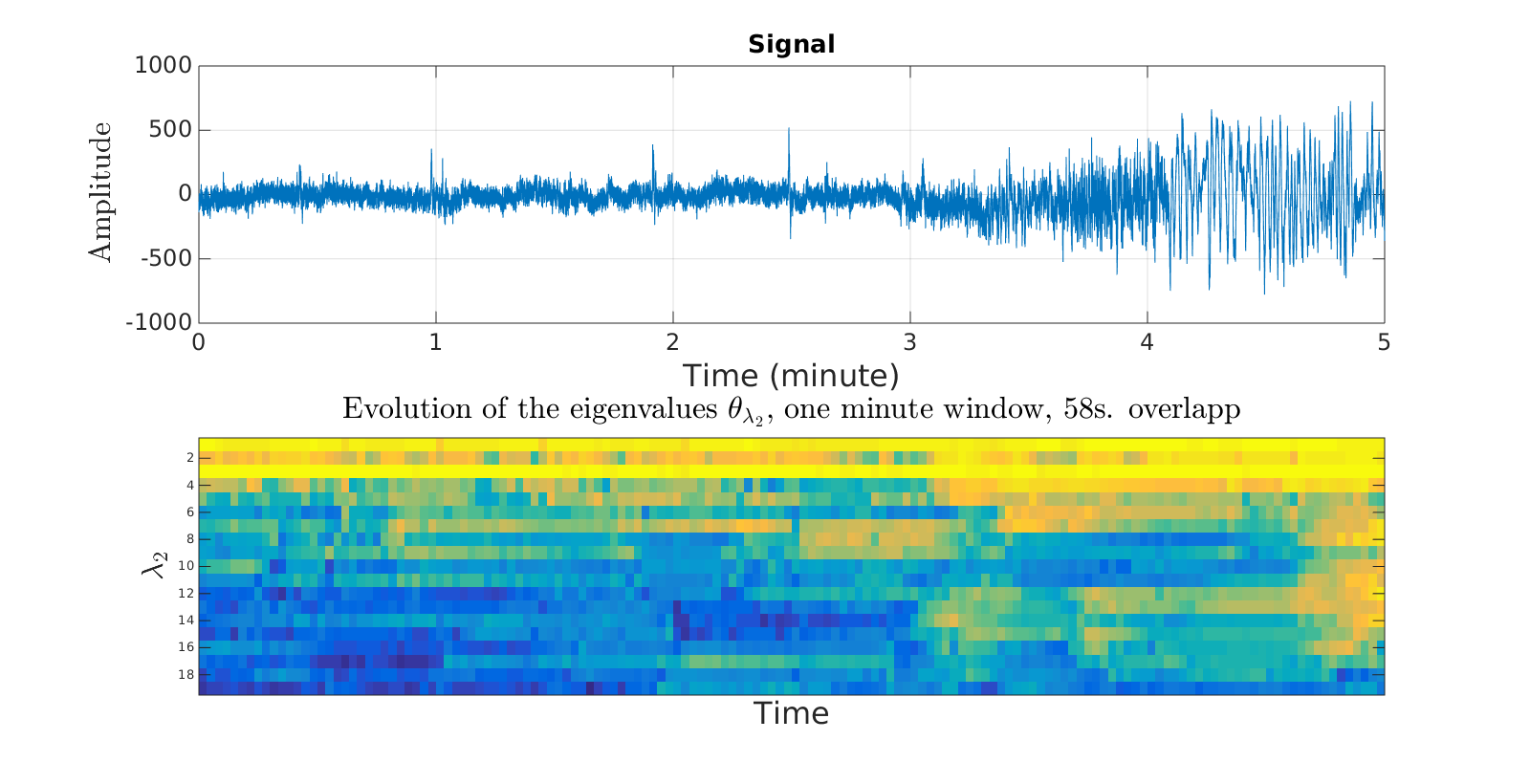}\caption{Evolution of the distribution of the eigenvalues over time highlighting the pre-ictal phase at time 3min.}\label{sparsity3}
\end{center}
\end{figure}
In the last case of Fig. \ref{sparsity3}, the change is more subtle around $3$ min. Taking a bigger window size or different parameters could lead to better representations or in this case, we might need information from $S_1$ to give better confidence in the pre-ictal hypothesis. In fact by looking at Fig. \ref{sparsity5} one can see the change in frequency representation. This is why all the coefficients should be kept ultimately.
As demonstrated, the eigenvalue distribution changes over time due to the appearance in the pre-ictal phase of clear frequency events leading to an efficient PCA implying in turn the change on the eigenvalues amplitude. We now demonstrate that the $Lx$ representation can also be used for transient detection.

\subsection{Inter-Ictal spike detection}

\begin{figure}[t!]
\begin{center}
\includegraphics[width=3.5in]{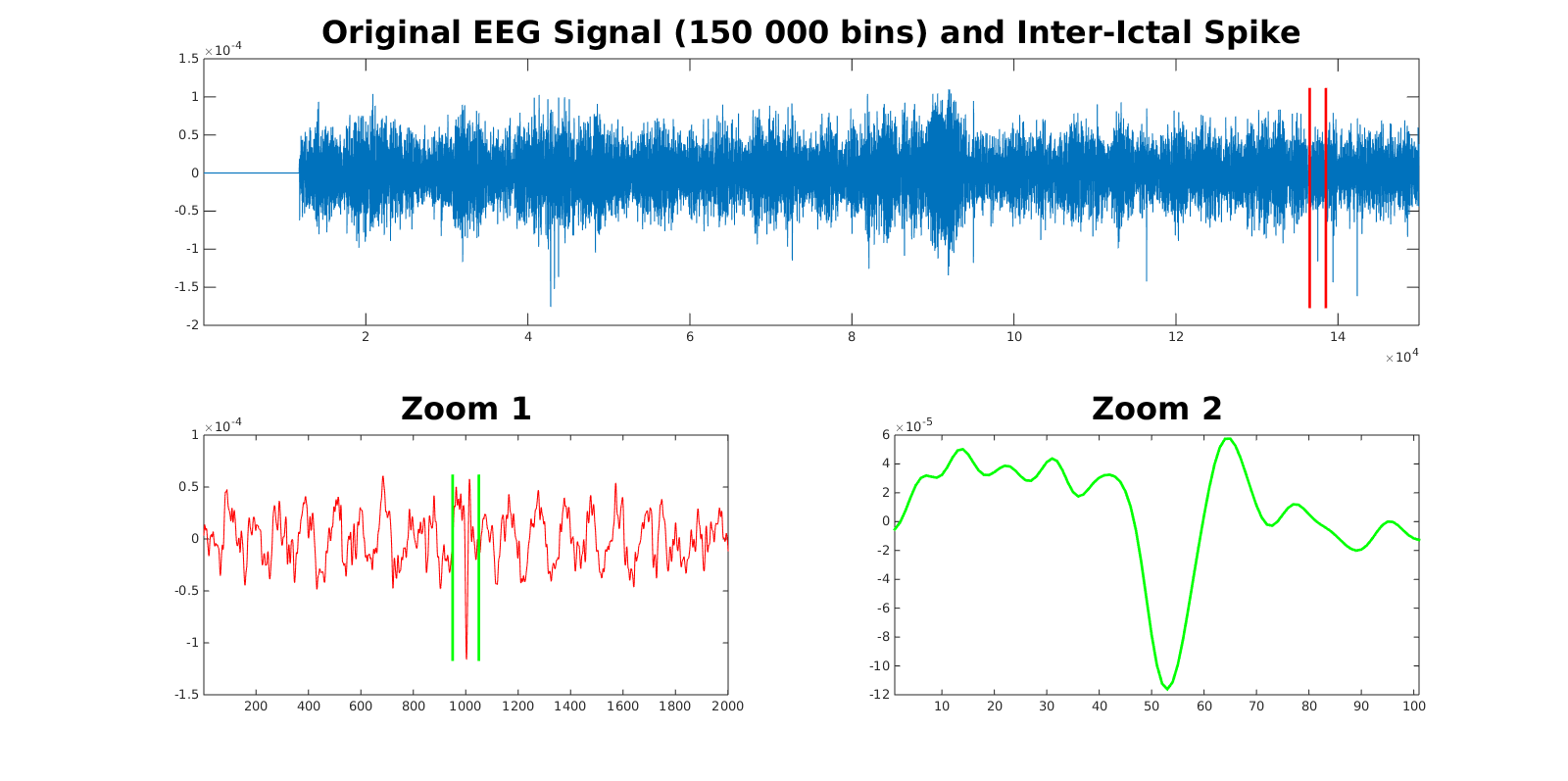}\caption{Treated EEG Signal and zoom on the feature of interest (the inter-ictal Spike)}\label{inter1}
\end{center}
\end{figure}

Another standard signal processing challenge is the one of transient detection which can be reformulated as a onset zone detection with the property that this zone is very localized in time and that the system returns to an ambient behavior.
Inter-ictal spike detection belongs to this class of problem, for which, we present a solution through the $Lx$ representation and the use of a clustering technique. If performed well, the dynamics of inter-ictal spikes can also provide meaningful information for tackling the seizure prediction task presented above and is therefor critical.
The signal and the features we are interested are presented in Fig. \ref{inter1}.
Firstly, one can notice that the SNR in $Lx$ is low due to the thresholding applying through $\rho$. In fact, when compared to the scattering coefficients, we are able to obtain a sparse representation with the transients only, without any background encoding.
We thus show that this nonlinearly separable problem of transient detection can be solved using our new representation.

\begin{figure}[t!]
\begin{center}
\includegraphics[width=3.8in]{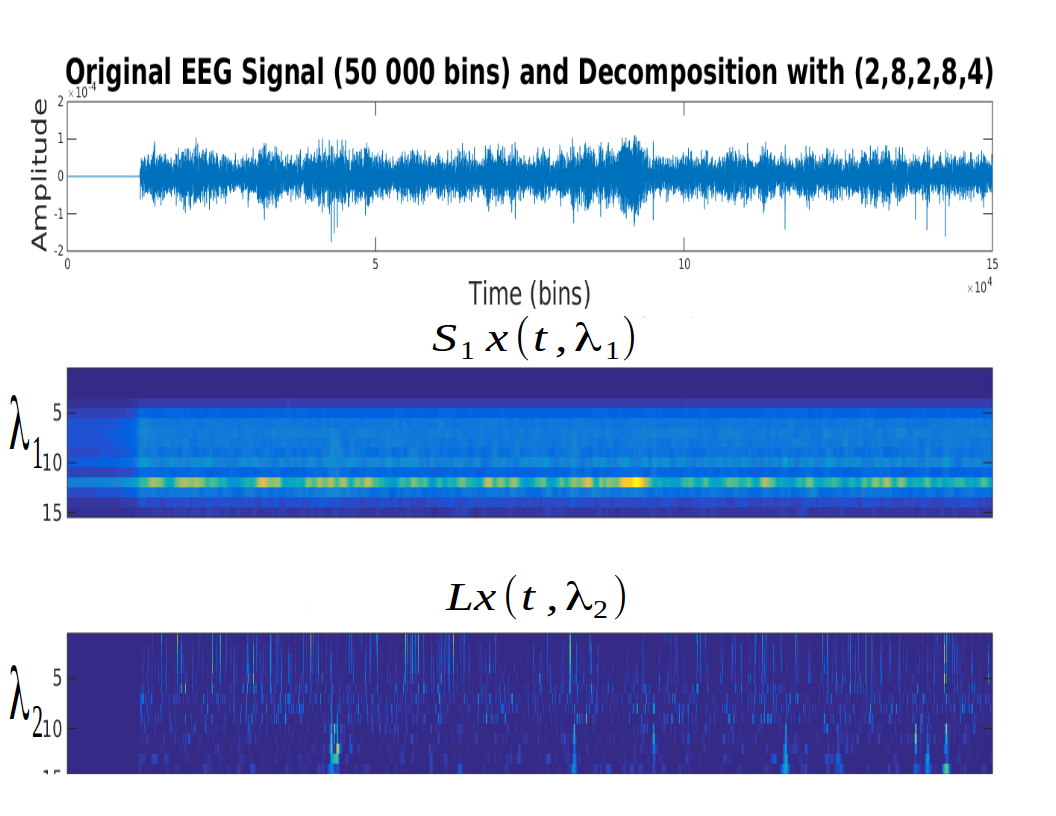}\caption{Treated signal, standard Scattering Coefficients $S_2x(t, \lambda_1)$ and $Lx(t,\lambda_2)$.}\label{inter2}
\end{center}
\end{figure}

As a qualitative measure we first show the sparsity of the representation in this noisy environment. Note that all the default parameters have been kept identical to the seizure prediction framework. In this case (Fig. \ref{inter2}) the raw scattering coefficients $S_2x$ are noisy and do not react to the actual transients (by construction) but capture the energy response of the signal to the used filters. 
With our representation , a new kind of information about transients is highlighted as seen in Fig. \ref{inter2}.
One can distinguish two clear clusters regarding the $\lambda_2$ dimension. It seems to show one kind of neural activity happening with small $\lambda_2$ (high frequency) and another cluster with slower transients. It seems that inter-ictal spikes belong to the latter.
In order to effectively detect and cluster the spikes without a priori knowledge we will perform an unsupervised clustering based on the $k$-NN clustering. 
This techniques will cluster observation based on pairwise distances evaluation and group together close points. However, it is necessary to first specify the number of clusters to use. Since we do not know a priori this quantity $k$, we resolve to cross-validate it with some qualitative measure of the clustering. We set $k \in \{ 1,...,6\}$ and apply this technique on the raw scattering first to show the absence of transient structures and then apply it on $Lx$ to analyze the improvement and the correctly clustered transients. We used the city-block distance in our experiment.

\paragraph{Unadapted Standard Scattering Coefficients}

\begin{figure}[t!]
\begin{center}
\includegraphics[width=3.5in]{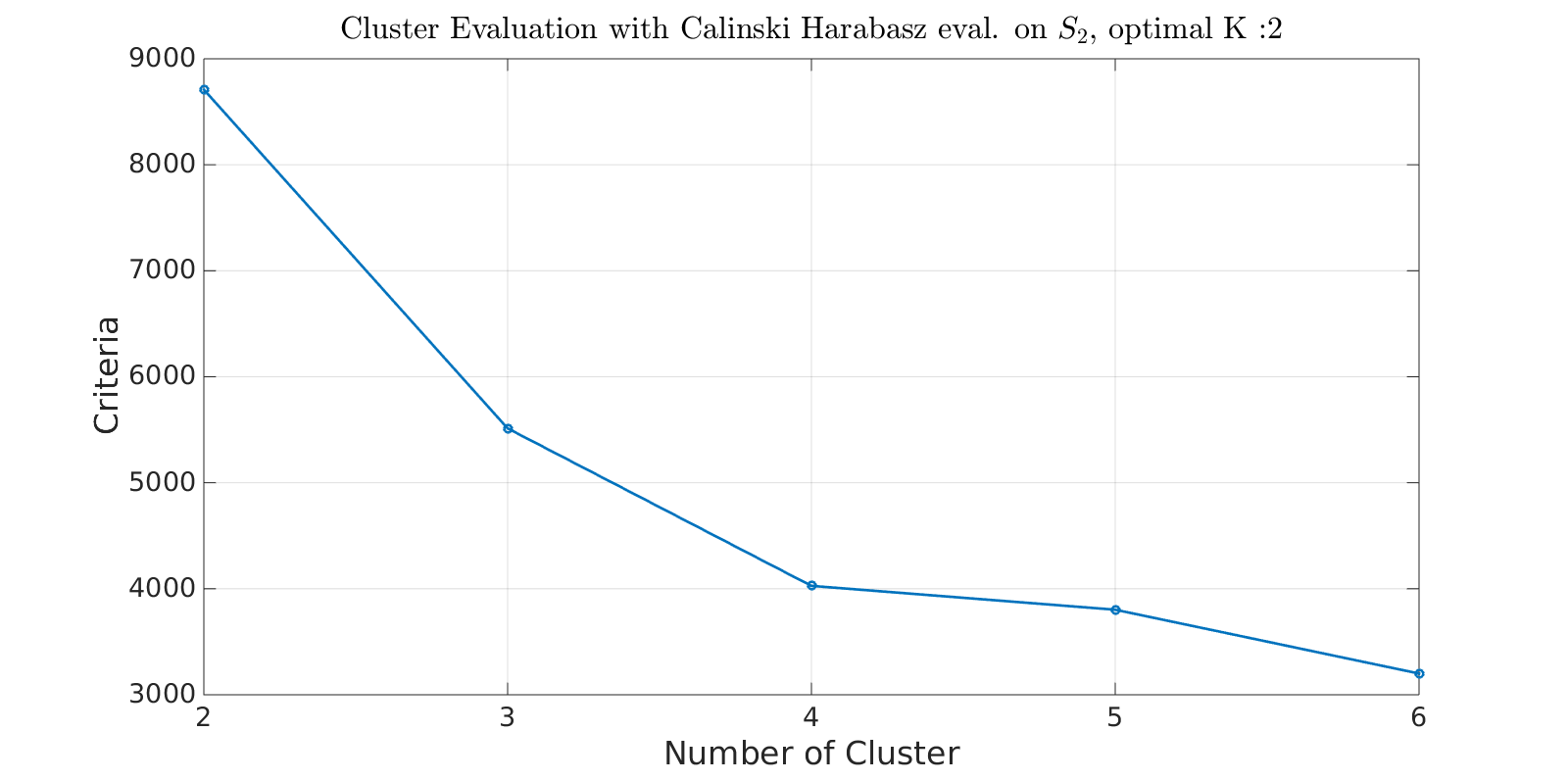}\caption{Optimal number of cluster selection based on a qualitative criteria.}\label{inter3}
\end{center}
\end{figure}

\begin{figure}[t!]
\begin{center}
\includegraphics[width=3.5in]{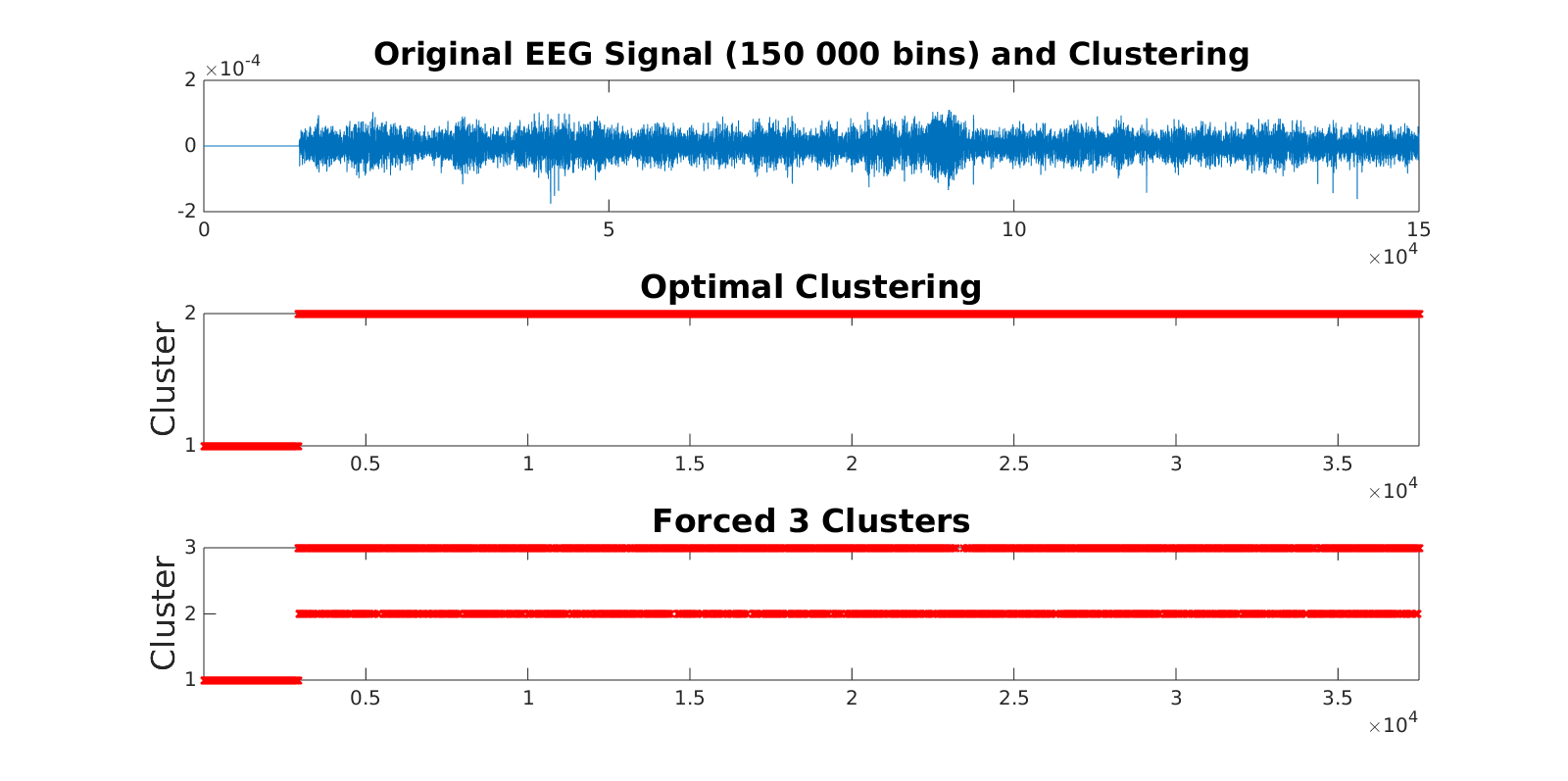}\caption{Treated Signal, clustering using $2$ clusters and $3$ clusters respectively.}\label{inter4}
\end{center}
\end{figure}

Using the standard Scattering Coefficients $Lx$, we found that the optimal number of cluster is $2$ (Fig. \ref{inter3}) thus we can now perform the clustering (Fig. \ref{inter4}).
Clearly the optimal clustering is based on the repartition energy over time and indeed perform the clustering : activity/silence. The transients however are not detected. In the case where we forced $k=3$, we can see that during the activity part of the signal, the two "activity" states alternate continuously between each other randomly. 

\begin{figure}[t!]
\begin{center}
\includegraphics[width=3.5in]{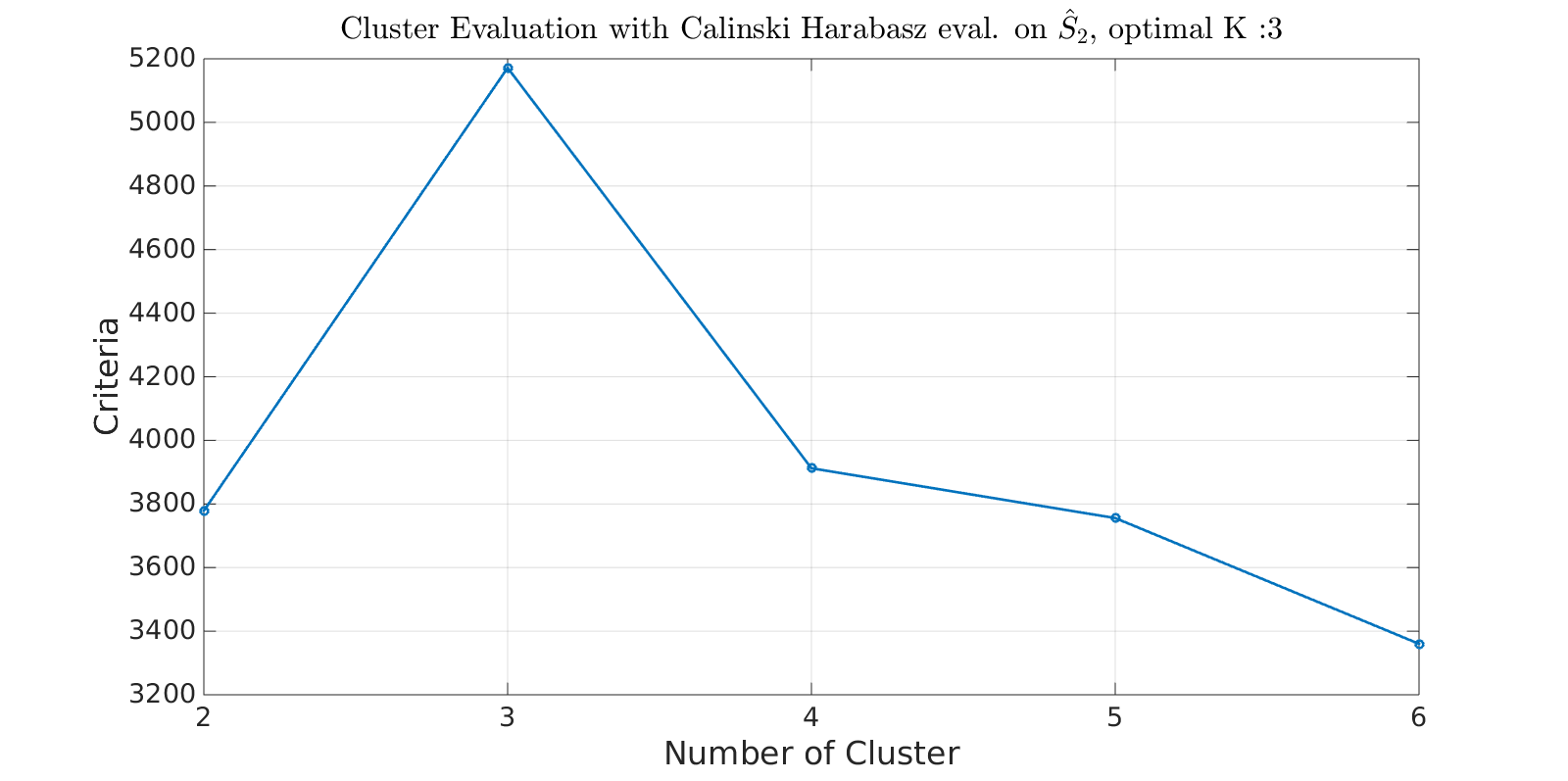}\caption{Optimal number of cluster selection based on a qualitative criteria ($k=3$)}\label{inter5}
\end{center}
\end{figure}

\begin{figure}[t!]
\begin{center}
\includegraphics[width=3.8in]{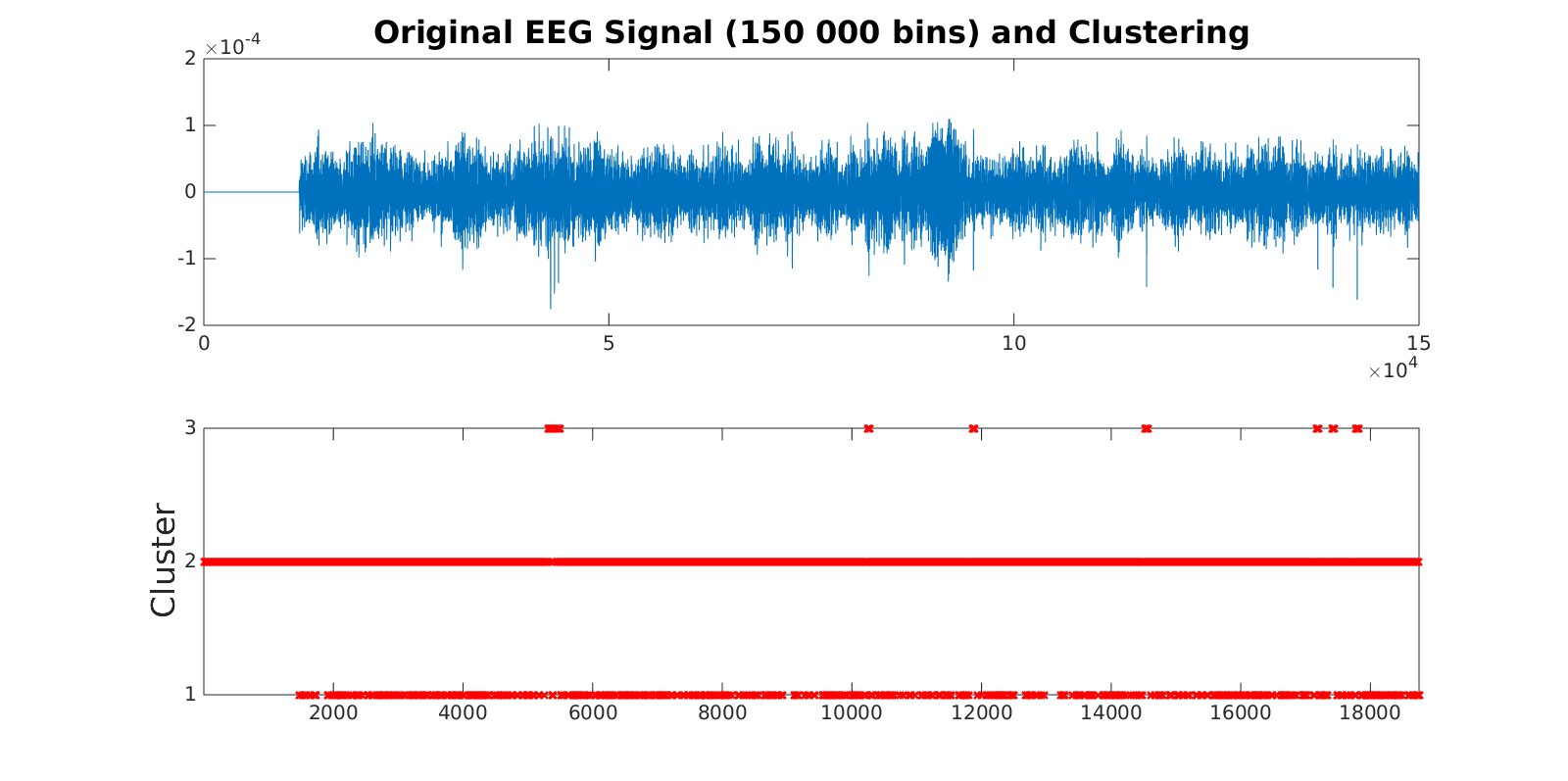}\caption{Treated signal and the optimal clustering (using $3$ clusters).}\label{inter6}
\end{center}
\end{figure}

\begin{figure}[t!]
\begin{center}
\includegraphics[width=3.8in]{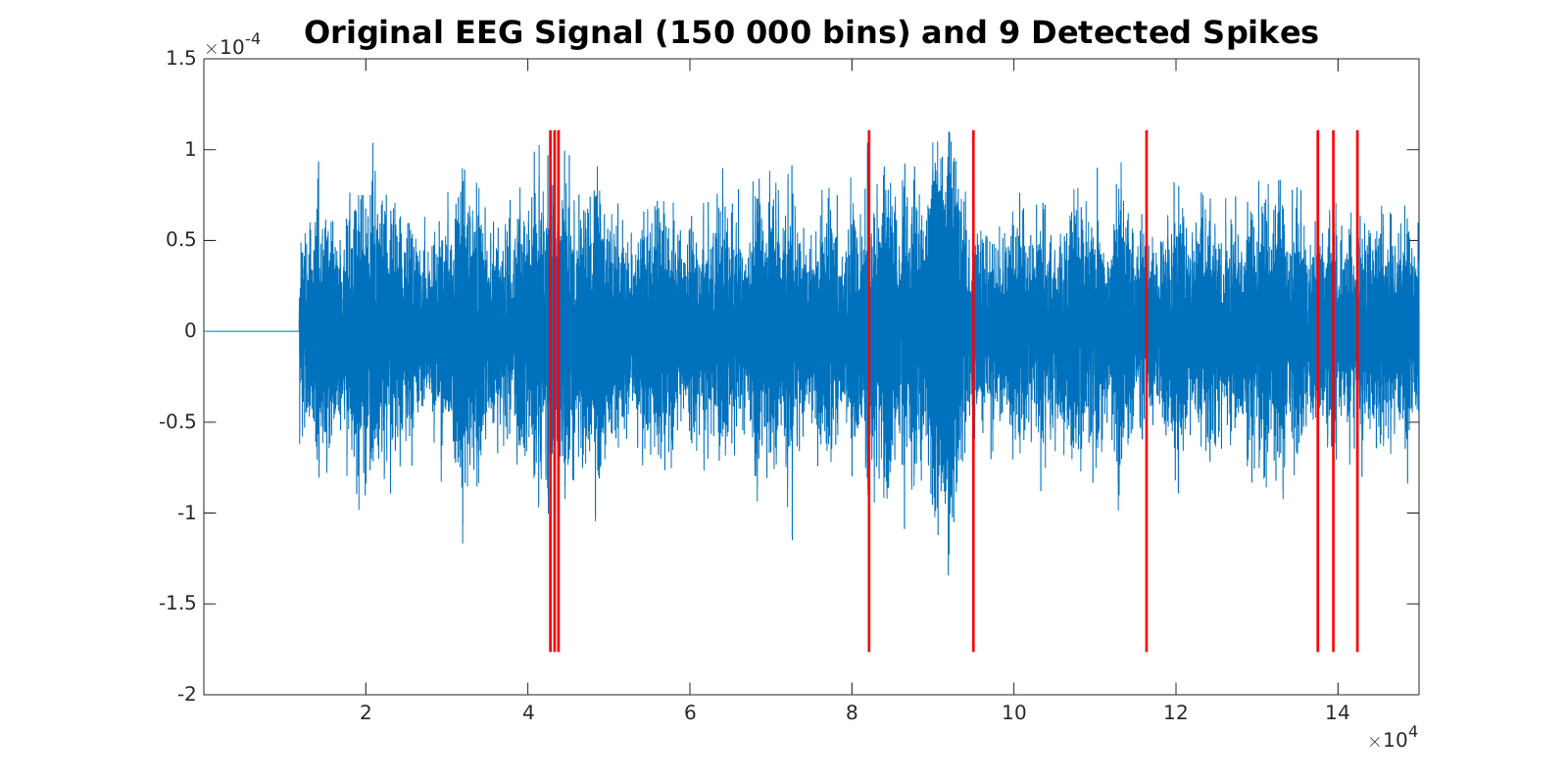}\caption{Treated Signal with the inter-ictal spikes detection in red}\label{inter7}
\end{center}
\end{figure}

\begin{figure}[t!]
\begin{center}
\includegraphics[width=3.8in]{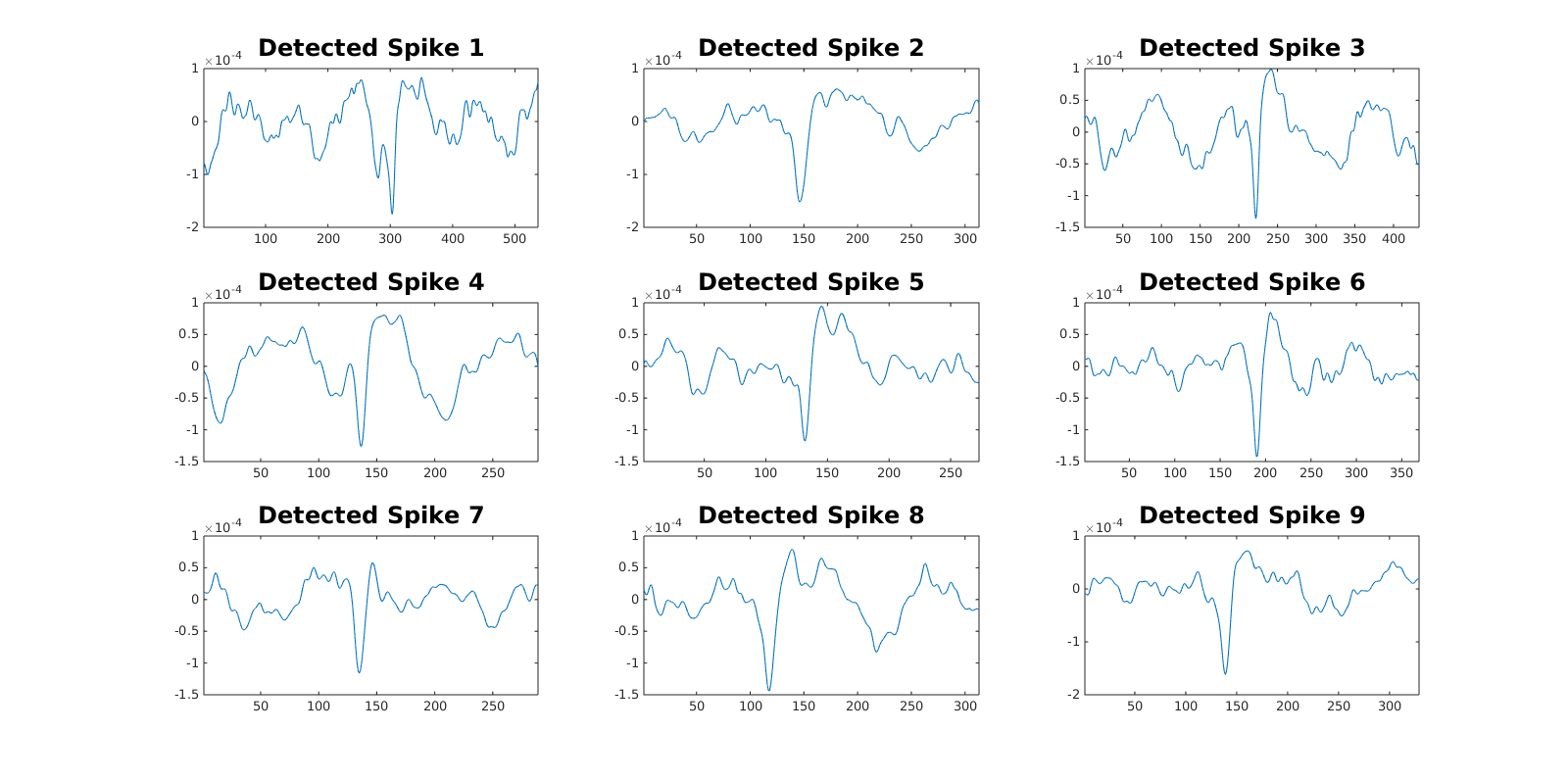}\caption{Detected spikes shown individually in a larger time window}\label{inter8}
\end{center}
\end{figure}

\paragraph{Efficiency of $Lx$}
On the other hand, when using $Lx$ as an input for our clustering algorithm, we found the optimal $k=3$ (Fig. \ref{inter5}). 
Given this optimal number of cluster, we perform once again the clustering (Fig. \ref{inter6}). This time, we end up with a clustering on $3$ states with the following possible interpretation. First, the normal brain activity will be denoted by $s_1$, the background noise as $s_2$ and finally inter-ictal spikes as $s_3$.
The result is robust to noise and to non transient activities.
Note that when forcing a clustering with $k=2$ we have an aggregation of $s_1$ and $s_2$, but $s_3$ is unchanged showing the clear cleavage between inter-ictal spikes and other activities when using this representation.
If we now look at the spikes position in the signal (Fig. \ref{inter7}) and the extracted spikes (Fig. \ref{inter8}) we can see the accuracy of this approach. Human analysis is matching this detection, there are no other inter-ictal spikes in the treated signal (no false negative) and each detected spike is indeed an inter-ictal spike (no false positive).

\section{Conclusion}
A new representation based on the scattering network was constructed and presented with the aim to highlight the transient structures present in a given signal. By construction, it is possible to discriminate different transient structures which in our case led to an optimal inter-ictal spike detection. The frequency invariance in addition of the time invariance property is a key feature of this new representation and should be considered as a key feature for this pre-processing technique. Even though the new representation by itself can't encode background information, when coupled with the standard scattering coefficients, it is possible to leverage both representations and have complementary information, about the stationary background and transients. We hope to further use this algorithm for different tasks where labeled data and bigger dataset are available in order to have a more quantitative way to express its usefulness and impact. On the other hand, a possible improvement is the use of Mixture of Probabilistic PCA instead of the standard PCA allowing nonlinear dimensionality reduction and possibly a better $Lx$ representations. Concerning the threshold, looking at a moving average or moving median instead of the median could bring more flexibility over time and better onset detections by not assuming a global stationary background but rather piecewise stationarity.
\section*{Acknowledgement}
I want to thank Mr. Vincent LOSTANLEN and Romain COSENTINO for their reviewing work and constructive comments and Mr. David RAMIREZ for his support.


\bibliographystyle{plain}

\bibliography{ref}

\begin{thebibliography}{10}

\bibitem{allard2014expansive}
Remy Allard and Jocelyn Faubert.
\newblock An expansive, cone-specific nonlinearity enabling the luminance
  motion system to process color-defined motion.
\newblock {\em Journal of vision}, 14(8):2--2, 2014.

\bibitem{anden2014deep}
Joakim And{\'e}n and St{\'e}phane Mallat.
\newblock Deep scattering spectrum.
\newblock {\em Signal Processing, IEEE Transactions on}, 62(16):4114--4128,
  2014.

\bibitem{randall3}
Randall Balestriero.
\newblock Scattering for bioacoustics.
\newblock Internship research report, Univ. Toulon, 2013-14 supervised by H.
  Glotin, 2013.

\bibitem{balestriero2015scattering}
Randall Balestriero et~al.
\newblock Scattering decomposition for massive signal classification: from
  theory to fast algorithm and implementation with validation on international
  bioacoustic benchmark.
\newblock In {\em 2015 IEEE International Conference on Data Mining Workshop
  (ICDMW)}, pages 753--761. IEEE, 2015.

\bibitem{randall4}
Randall Balestriero and Herve Glotin.
\newblock Humpback whale song representation.
\newblock NIPS4B 2013, 2013.

\bibitem{bruna2013invariant}
Joan Bruna and St{\'e}phane Mallat.
\newblock Invariant scattering convolution networks.
\newblock {\em Pattern Analysis and Machine Intelligence, IEEE Transactions
  on}, 35(8):1872--1886, 2013.

\bibitem{chen2013music}
Xu~Chen and Peter~J Ramadge.
\newblock Music genre classification using multiscale scattering and sparse
  representations.
\newblock In {\em Information Sciences and Systems (CISS), 2013 47th Annual
  Conference on}, pages 1--6. IEEE, 2013.

\bibitem{cortes1995support}
Corinna Cortes and Vladimir Vapnik.
\newblock Support vector machine.
\newblock {\em Machine learning}, 20(3):273--297, 1995.

\bibitem{m3}
DI~ENS.
\newblock Scatnet toolbox.
\newblock http://www.di.ens.fr/data/software/scatnet/, 2011-2015.

\bibitem{giusti2013fast}
Alessandro Giusti, Dan~C Cire{\c{s}}an, Jonathan Masci, Luca~M Gambardella, and
  J{\"u}rgen Schmidhuber.
\newblock Fast image scanning with deep max-pooling convolutional neural
  networks.
\newblock {\em arXiv preprint arXiv:1302.1700}, 2013.

\bibitem{randall5}
H~Goeau, H~Glotin, WP~Vellinga, and A~Rauber.
\newblock Lifeclef bird identitfication task 2014.
\newblock Clef working notes, 2014.

\bibitem{guirgis2013role}
Mirna Guirgis, Yotin Chinvarun, Peter~L Carlen, and Berj~L Bardakjian.
\newblock The role of delta-modulated high frequency oscillations in seizure
  state classification.
\newblock In {\em Engineering in Medicine and Biology Society (EMBC), 2013 35th
  Annual International Conference of the IEEE}, pages 6595--6598. IEEE, 2013.

\bibitem{m2}
Anden J. and Mallat S.
\newblock Deep scattering spectrum.
\newblock Deep Scattering Spectrum,Submitted to IEEE Transactions on Signal
  Processing, 2011.

\bibitem{PCA}
I.~T. Jolliffe.
\newblock {\em Principal Component Analysis}.
\newblock Springer, 2002.

\bibitem{mallat1999}
Alexis Joly, Herve Goeau, Herve Glotin, Concetto Spampinato, and Henning
  Muller.
\newblock {\em Lifeclef 2014: multimedia life species identification
  challenges}.
\newblock Information Access Evaluation. Multilinguality, Multimodality, and
  Interaction, Springer International Publishing, 2014.

\bibitem{lecun1995convolutional}
Yann LeCun and Yoshua Bengio.
\newblock Convolutional networks for images, speech, and time series.
\newblock {\em The handbook of brain theory and neural networks}, 3361(10),
  1995.

\bibitem{mallat1999wavelet}
St{\'e}phane Mallat.
\newblock {\em A wavelet tour of signal processing}.
\newblock Academic press, 1999.

\bibitem{m1}
Stephane MALLAT.
\newblock Group invariant scattering.
\newblock Communications in Pure and Applied Mathematics, vol. 65, no. 10, pp.
  1331-1398, 2012.

\bibitem{meyer1993wavelets}
Yves Meyer.
\newblock Wavelets-algorithms and applications.
\newblock {\em Wavelets-Algorithms and applications Society for Industrial and
  Applied Mathematics Translation., 142 p.}, 1, 1993.

\bibitem{nagi2011max}
Jawad Nagi, Frederick Ducatelle, Gianni~A Di~Caro, Dan Cire{\c{s}}an, Ueli
  Meier, Alessandro Giusti, Farrukh Nagi, J{\"u}rgen Schmidhuber, and
  Luca~Maria Gambardella.
\newblock Max-pooling convolutional neural networks for vision-based hand
  gesture recognition.
\newblock In {\em Signal and Image Processing Applications (ICSIPA), 2011 IEEE
  International Conference on}, pages 342--347. IEEE, 2011.

\bibitem{oyallon2015deep}
Edouard Oyallon and St{\'e}phane Mallat.
\newblock Deep roto-translation scattering for object classification.
\newblock In {\em Proceedings of the IEEE Conference on Computer Vision and
  Pattern Recognition}, pages 2865--2873, 2015.

\bibitem{quiroga2002frequency}
R~Quian Quiroga, H~Garcia, and A~Rabinowicz.
\newblock Frequency evolution during tonic-clonic seizures.
\newblock {\em Electromyography and clinical neurophysiology}, 42(6):323--332,
  2002.

\bibitem{reed1908methods}
Michael Reed and Barry Simon.
\newblock {\em Methods of modern mathematical physics}, volume~2.
\newblock Academic press, 1908.

\bibitem{sifre2013rotation}
Laurent Sifre and St{\'e}phane Mallat.
\newblock Rotation, scaling and deformation invariant scattering for texture
  discrimination.
\newblock In {\em Proceedings of the IEEE Conference on Computer Vision and
  Pattern Recognition}, pages 1233--1240, 2013.

\bibitem{Sifre_2013_CVPR}
Laurent Sifre and Stephane Mallat.
\newblock Rotation, scaling and deformation invariant scattering for texture
  discrimination.
\newblock In {\em The IEEE Conference on Computer Vision and Pattern
  Recognition (CVPR)}, June 2013.

\bibitem{tipping1999probabilistic}
Michael~E Tipping and Christopher~M Bishop.
\newblock Probabilistic principal component analysis.
\newblock {\em Journal of the Royal Statistical Society: Series B (Statistical
  Methodology)}, 61(3):611--622, 1999.

\bibitem{randall2}
Trone, Balestriero, and Glotin.
\newblock Gabor scalogram reveals formants in high-frequency dolphin clicks.
\newblock Proc. of Neural Information Processing Scaled for Bioacoustics: from
  Neurons to Big Data, 2013, Ed. Glotin H., LeCun Y., Artieres T., Mallat S.,
  Tchernichovski O., Halkias X., joint to NIPS Conf.,
  http://sabiod.org/publications.html, ISSN 979-10-90821-04-0, 2013.

\bibitem{randall1}
Marie Trone, Herve Glotin, Randall Balestriero, and Bonnett~E David.
\newblock All clicks are not created equally: Variations in high-frequency
  acoustic signal parameters of the amazon river dolphin (inia geoffrensis).
\newblock in The Journal of the Acoustical Society of America, Volume 136,
  Issue 4, short letter, Oct 2014, long paper in preparation., 2014.

\end{thebibliography}
\nocite{*}


\end{document}